\documentclass[oneside,english]{amsart}
\usepackage[T1]{fontenc}
\usepackage[latin9]{inputenc}
\usepackage{float}
\usepackage{amsthm}
\usepackage{amsbsy}
\usepackage{graphicx}
\usepackage{amssymb}
\usepackage{esint}\usepackage{color}
\usepackage{epstopdf}

\makeatletter

\newcommand{\lyxdot}{.}

\floatstyle{ruled}
\newfloat{algorithm}{tbp}{loa}
\floatname{algorithm}{Algorithm}

\usepackage{bbm}

\makeatother

\usepackage{babel}

\begin{document}

\title[Oscillatory-Network learning]{A Bayesian  approach for structure learning in oscillating regulatory networks}
\author[D Trejo, AJ Millar and G Sanguinetti ]{Daniel Trejo Banos\,$^{1}$, Andrew J. Millar\,$^{2,3}$ and Guido Sanguinetti\,$^{1,2,}$\\
\\
$^{1}$School of Informatics, University of Edinburgh, 10 Crichton St, Edinburgh EH8 9AB, UK.\\
$^{2}$SynthSys - Systems and Synthetic Biology, University of Edinburgh, CH Waddington Building, King's Buildings, Mayfield Road, Edinburgh EH9 3JD, UK.\\
$^{3}$School of Biological Sciences, University of Edinburgh, Darwin Building, King's Buildings, Mayfield Road, Edinburgh EH9 3JR, UK.}

\maketitle

\begin{abstract}

Oscillations lie at the core of many biological processes, from the cell cycle, to circadian oscillations and developmental processes. Time-keeping mechanisms are essential to enable organisms
to adapt to varying conditions in environmental
cycles, from day/night to seasonal. Transcriptional regulatory networks are
one of the mechanisms behind these biological oscillations. However, while identifying cyclically expressed genes from time series measurements is relatively easy, determining the structure of the interaction network underpinning the oscillation is a far more challenging problem.
Here, we explicitly leverage the oscillatory nature of the transcriptional signals and present a method  for reconstructing network interactions tailored to this special but important class of genetic circuits. Our method is based on projecting the signal onto a set of oscillatory basis functions using a Discrete Fourier Transform. We build
a Bayesian Hierarchical model within a frequency domain linear
model in order to enforce sparsity and incorporate prior
knowledge about the network structure.  Experiments on real and simulated data show that the method can lead to substantial improvements over competing approaches if the oscillatory assumption is met, and remains competitive also in cases it is not.
\section{Availability:}
DSS, experiment scripts and data are available at
http://homepages.inf.ed.ac.uk/gsanguin/DSS.zip
\section{Contact:}
D.Trejo-Banos@sms.ed.ac.uk
\end{abstract}

\section{Introduction}
Cyclic behaviour is ubiquitous in biology. The importance of oscillatory systems stems both from the necessity to adapt to the many environmental cycles (circadian, annual, etc.), as well as to maintain intrinsically periodic processes such as the cell cycle. Both of these type of oscillations are essential to many physiological processes, and malfunctions in the cellular time keeping mechanisms are frequently associated with disease, further motivating the study of these systems \cite{bell_2005}.

 Genetic regulatory networks
are at the core of many of these biological oscillators. These networks can sustain oscillatory behaviour  in protein levels through specific architectures involving multiple feedback loops of transcriptional regulation. For example, a transcriptional
oscillator is thought to drive the \emph{Arabidopsis thaliana} circadian
clock through mutual repression of three transcriptional regulators  \cite{pokhilko_clock_2012,mcclung_chapter_2011}. The cell cycle is another oscillatory process, which controls cell
division and duplication. In the case of \emph{Saccharomyces cerevisiae},
experiments and dynamical models suggest that the cell cycle is the
result of a transition between two self maintaining steady states, driven
by two antagonistic classes of proteins \cite{chen_integrative_2004}.
Evidence suggests that a transcriptional network is an important part
of this mechanism \cite{spellman_comprehensive_1998,li_yeast_2004,orlando_global_2008}.

These oscillators have been the subject of study for many years, but
uncovering the exact mechanism is a challenge that involve many complex
chemical, genetic and physiological components. It is therefore important to devise computational statistical methods which may guide experimental analyses by inferring potential regulatory interactions directly from time series gene expression data, which is usually easier to obtain. 

Network inference is a well established and rich domain of research in systems biology. State of the
art methods for regulatory network inference include a wide variety
of techniques from statistics and machine learning. For example,
mutual information between gene expression levels under different
experimental conditions is used by ARACNE \cite{margolin_aracne:_2006}
and CLR \cite{faith_large-scale_2007}, two of the most widely used methods for network reconstruction. GENIE3 \cite{huynh-thu_inferring_2010}, another method which was a top performer at the DREAM network inference challenges, uses random forests
to produce a weighted ranking over the network edges. Other methods recently used include
regularized regression \cite{haury_tigress:_2012}, ANOVA
\cite{kuffner_inferring_2012} and Hierarchical Gaussian models \cite{li_inferring_2006}
Most of these methods focus on steady state data, which is by definition not available for oscillatory networks.

Regularisation-based and Bayesian methods can also be adapted to time series data. Dynamic Bayesian Networks have long been a popular choice in network inference (e.g. \cite{oates_network_2012}). Such methods present considerable advantages in being able to quantify uncertainty and to incorporate prior knowledge, but are often severely limited by computational constraints. Optimisation-based methods based on regularised regression \cite{bonneau_inferelator:_2006} present often a scalable alternative at the cost however of some modelling flexibility. 

Here, we use a first order model of the system dynamics
to constrain the network inference, but we explicitly take advantage of the oscillatory behaviour of the system by pursuing frequency-based estimation. We build
a hierarchical Bayesian model over the network dynamics which can set
and infer structural constraints and account for the inevitable uncertainty
that experimental settings convey. Furthermore, our method can easily integrate non-trivial side information, for example in the form of sequence similarity between promoter sequence of genes. Experimental results on real and simulated data highlight that the method offers an effective and flexible platform for statistical inference in oscillatory systems, and can uncover non-trivial biological information.

The rest of the paper is organised as follows: the next section describes the methodology we use, reviewing the linear time-invariant approximation we use as well as introducing the Bayesian hierarchical framework for network inference. We then present an experimental evaluation on three data sets: a synthetic data set from the DREAM network inference challenge, a simulated data set obtained  from a state of the art model of the {\it A. thaliana} circadian clock  \cite{pokhilko_data_2010}, and a real data set from the yeast {\it S. cerevisiae} cell cycle \cite{orlando_global_2008}. We then conclude the paper by discussing our method in the light of these experimental results and the existing literature on network inference.

\section{Methods}
Our approach is centred on the assumption that the oscillatory dynamics of the regulatory network can be reasonably approximated, in Fourier space, by a linear time invariant system. This is of course a simplification, but it is not an unreasonable one, and has been previously proposed as a formalism to model oscillatory genetic circuits with considerable success, see \cite{dalchau_understanding_2012} for a recent review. From the inferential point of view, adopting a frequency domain perspective is convenient, as it enables us to transform the network reconstruction problem in a regression problem, for which many advanced estimation tools exist. We choose a Bayesian regression approach, as it provides an effective methodology to integrate diverse information in the inferential machine. As a proof of principle of how non-trivial information can be incorporated, we discuss how sequence similarity between promoter regions could be used within a hierarchical model framework.
\subsection{Linear time invariant model}

The starting point for our modelling is the approximation of the system's dynamics as a Linear Time Invariant (LTI) model:
\begin{equation}
{\frac{dx_i\left(t\right)}{dt}=\sum_{j\neq i}^{N}\alpha_{ij}x_j\left(t\right)}+{{b}_{i}}-\lambda_{i}{x_i\left(t\right)}+\sum_{k}{c_{ik}}{{u}_{k}}.\label{eq:aproxi}\end{equation}

Here the expression level of gene ${i}$, denoted as ${x_i\left(t\right)}$, depends
on the expression levels of the other ${N-1}$ genes (potential regulators) through activating or repressing intensity $\alpha_{ij}\in\mathbb{R}$.
Gene expression levels decay linearly with rates $\lambda_{i}$. Additionally, gene expression depends on a set of $K$ inputs ${u}_{k}$ which can be either external
signals (light for example) or any other gene signal that is not modelled explicitly in the network. Finally, each gene has a basal transcription rate  ${b}_{i}$.

Having a set of ${M}$ samples from an experiment (e.g. mRNA levels
from a microarray experiment), let the vector $\mathbf{x}_{i}\in\mathbb{R^{M}}$
denote the set of ${M}$ expression level measurements for
gene ${i}$. We can further construct the matrix $X\in\mathbb{R}^{{M}\times{N}}$,
which contains the sample points for the set of ${N}$ genes.
Let $\dot{X}$ be the derivative of $X$,
so equation (\ref{eq:aproxi}) in matrix form for this set of gene
expression levels is given by:

\begin{equation}
\mathrm{\dot{X}=}X{A^{T}}+\mathbf{b\mathbf{1}}+U{C^{T}}\label{eq:matrix form}\end{equation}
where ${A}\in\mathbb{R}^{{N}\times{N}}$ is the
matrix with diagonal elements $\lambda_{i}$ and off-diagonal elements
$\alpha_{ij}$, the input signals are contained in matrix $U\in\mathbb{R}^{M\times K}$. To complete the notation, we denote with $\mathbf{b}$ vector of basal expression levels, which multiplies the ${M}\times{N}$ matrix of ones $\mathbf{1}$ to add a constant term to the equation.

We proceed to compute the derivative $\dot{\mathbf{x}}$ by first
projecting the gene expression levels into a set of orthogonal basis
functions. The chosen set of basis functions is the one given by the
Discrete Fourier Transform of the gene expression levels. We emphasize that the choice of basis function is dictated by the nature of the problem: while in the limit of a continuously sampled signal this choice would be irrelevant (any complete basis would yield perfect reconstruction), for discretely sampled signals the quality of the approximation to the signal (and its derivative) will depend on the expressiveness of the chosen finite set of basis functions. Our choice of basis functions is motivated by the prior knowledge that the signals of interest should be oscillatory, making the choice to work in the frequency domain particularly appealing. We denote
$\mathbf{X\left(\boldsymbol{\omega}\right)}$, $\mathbf{X}$ for brevity, as the frequency representation of $\mathbf{x}$, with
each column containing the frequency spectrum of the expression of a gene over the time points.  The frequency domain derivative  can be computed
analytically by $\mathbf{\dot{X}}=2\pi\boldsymbol{\omega} i \mathbf{X}$, so the frequency domain representation of the system
is given by:

\begin{equation}
\mathrm{\dot{\mathbf{X}}=}\mathbf{X}\mathbf{A^{T}}+\mathbf{U}\mathbf{C^{T}}.\label{eq:matrix form-1}\end{equation}
Basal rates $\mathbf{b}$ are included in the zero frequency
component of $\mathbf{X}$. The frequency representation of the inputs
is given by $\mathbf{U}$.

To account for any discrepancies between the linearised model and
the true system dynamics, we assume normally distributed error with
variance $\sigma_{D}^2$. The likelihood function for equation (\ref{eq:matrix form-1}) is:

\begin{equation}
\begin{gathered}\mathrm{p}\left(\mathbf{\dot{X}|}\mathbf{X},\mathbf{A},\mathbf{U},\mathbf{C},\sigma_{D}\right)\propto\ \prod_{i=1}^{\mathrm{N}} \sigma_{D}^{\mathrm{-M}}\exp\left(-\frac{1}{2\sigma_{D}^{2}}\mathbf{Q_{i}}\right)\\
\mathbf{Q_{i}}=\left(\mathrm{\dot{\mathbf{X_{i}}}-}[\begin{array}{cc}
\mathbf{X} & \mathbf{U}\end{array}]\left[\mathbf{\begin{array}{c}
\mathbf{A_{i}^{T}}\\
\mathbf{C}_{i}^{T}\end{array}}\right]\right)^{\mathbf{T}}\mathbf{\left(\mathrm{\dot{\mathbf{X_{i}}}-}[\begin{array}{cc}
\mathbf{X} & \mathbf{U}\end{array}]\left[\mathbf{\begin{array}{c}
\mathbf{A_{i}^{T}}\\
\mathbf{C}_{i}^{T}\end{array}}\right]\right)}\end{gathered}
.\label{eq:likelihood}\end{equation}

In general, multiple replicate time series may be available. Denoting with ${K}$ the number of replicate time series,
the overall likelihood, under an assumption of normal i.i.d error between series, can be generalized as:

\begin{equation}
\mathrm{P}\left(\left\{ \dot{\mathbf{X}}_{k}\right\} |\left\{ \mathbf{X}_{k}\right\} \mathbf{A,U,C},\sigma_{D}\right)=\prod_{k=1}^{\mathrm{K}}\mathrm{P}\left(\dot{\mathbf{X}}_{k}|\mathbf{X}_{k},\mathbf{A,U,C},\sigma_{D}\right)
\label{eq:product}
\end{equation}

which is a product of Gaussian densities.

Notice that the form of equation \eqref{eq:product} is identical to a regression problem where the output variables (Fourier coefficients of the derivatives of the signals) are regressed onto the Fourier coefficients of the signals. The inference problem of estimating the interaction and input response matrices $\left[\begin{array}{cc}
\mathbf{A}^{\mathbf{T}} & \mathbf{C^{T}}\end{array}\right]^{\mathrm{T}}$ in equation (\ref{eq:likelihood}) can therefore be attacked using the vast repertoire of regression methods.
Regularized regression methods have been tested in a network inference
context, see \cite{charbonnier_weighted-lasso_2010,bergersen_weighted_2011,bonneau_inferelator:_2006,haury_tigress:_2012}. Here, we opt for a hierarchical Bayesian
approach, that will allow us to leverage prior knowledge and integrate
other sources of information.

\subsection{Hierarchical Bayesian modelling}

To interpret dynamical systems in a network perspective, we assume that the interaction matrix in our LTI representation \eqref{eq:aproxi} has a sparse structure representing discrete interactions between regulators and target genes.
We introduce the {\it structural adjacency matrix} $\mathbf{H}\in\mathbb{R}^{\mathrm{N}\times\mathrm{N}}$, which sits at the top of the hierarchy. This matrix contains elements $h_{ij}=1$
if gene $j$ regulates gene $i$ for $i\neq j$. In this Bayesian approach, a sparsity
inducing prior over elements of $\mathbf{H}$ is necessary to aid identifiability and interpret-ability. The prior form chosen for elements $h_{ij}$ is a Bernoulli
distribution, with parameter $w$ which has a Beta distribution prior due to conjugacy.

We chose a spike and slab prior to relate the connection matrix $\mathbf{H}$
and interaction matrix $\mathbf{A}$. This distribution consists of
a mixture of a degenerate distribution and a long tailed distribution.
The form chosen is derived from the one presented in \cite{ishwaran_spike_2005},
where the $a_{ij}$ elements are drawn from a scale-mixture model where a zero-mean normal distribution
has variance governed by hyper-parameter $\tau_{ij}$. In this form,
the hyper-variance $\mathrm{h}_{ij}\tau_{ij}^{2}$ has a continuous
bi-modal distribution. With this prior, the posterior distribution of the less relevant parameters is shrunk towards zero and the non-zero elements are selected by the distributions tail. The advantage of the continuous  distribution implied by the scale-mixture model of \cite{ishwaran_spike_2005} lies primarily in the fact that we avoid the need to parametrize these bimodal distributions manually.

Thus, the hierarchical model is defined by equations:

\begin{equation}
\begin{array}{ccc}
{\mathrm{P}\left(\left\{ \dot{\mathbf{X}}_{k}\right\} |\left\{ \mathbf{X}_{k}\right\} \mathbf{A,C},\mathbf{U},\sigma_{D}\right)} & = & {\prod_{k=1}^{\mathrm{K}}\mathrm{P}\left(\dot{\mathbf{X}}_{k}|\mathbf{A,C},\mathbf{U},\mathbf{X}_{k},\sigma_{D}\right)}\\
{\mathrm{P}\left(a_{ij}|h_{ij},\tau_{ij}\right)} & \sim & \mathcal{N}\left(0,h_{ij}\tau_{ij}^{2}\right)\\
{\mathrm{P}\left(h_{ij}|w\right)} & \sim & { \left(1-w\right)\delta_{v0}+w\delta_{1}}\\
{\pi\left(w\right)} & \sim & { \mathrm{Beta}\left(\mathrm{a}_{1},\mathrm{a}_{2}\right)}\\
{\pi\left(\tau^{-2}\right)} & \sim & { \mathrm{Gamma\left(b_{1},\mathrm{b}_{2}\right)}}\\
{\pi\left(\sigma_{D}^{-2}\right)} & \sim & { \mathrm{Gamma}\left(\mathrm{c_{1},\mathrm{c}_{2}}\right)}.\end{array}\label{eq:model}\end{equation}

The parameter $\sigma_{D}$ accounts for uncertainty related to noise
and model mismatch, for example arising from the linear approximation to the system
dynamics. The parameter $v0$ is introduced for numerical stability and is fixed to the value of 0.005. The hyperparameters $a_{1,2}$, $b_{1,2}$ and $c_{1,2}$ can be fixed to reflect prior beliefs, or set to vague values to reflect prior ignorance; in the rest of the paper they are set to the default values of (1, 1), (5 , 50) and (0.001,0.001) respectively.

\subsection{Sequence information integration}

A major advantage of hierarchical modelling is the possibility
of integrating different data sources. By branching from the top of
the hierarchy, we can define models for different network related
characteristics and keep all the information coupled by the top of
the hierarchy. For example,  protein interaction and binding
data from ChIP-chip or ChIP-seq experiments can be used in a straightforward manner to modulate the
prior probabilities over matrix $\mathbf{H}$, for example by adjusting the parameter $w$ for individual edges. 

Hierarchical models also allow us to exploit more subtle sources
of structural information derived from an analysis of sequence information. Transcription factors bind to the promoter region of their targets by recognizing specific motifs, short DNA words; thus
co-regulated genes (genes that are regulated by a common transcription factor) should share common motifs in their promoted regions. We
use this information to draw the basic model for our sequence integration
approach. As the transcription binding sites share a common motif,
we assume that the similarity between two promoter regions varies
proportionally to the number of shared regulators. In this way, an observed pairwise similarity
matrix $\mathbf{S}=[s_{ij}]$ between gene promoters, derived from a multiple alignment method like  \cite{sievers_fast_2011}
or an alignment-free method \cite{Bonham-Carter31072013}, can be related to the structural adjacency matrix at the top of the hierarchical model. Assuming for simplicity a Gaussian observation model, we can then incorporate sequence similarity by positing the following relationship between promoter similarity scores and the structural adjacency matrix

\begin{equation}
\mathrm{p}\left(s_{ij}|\mathbf{H},\boldsymbol{\beta},\sigma_{seq}\right)\propto\sigma_{seq}^{-1/2}\exp\left(-\frac{1}{2\sigma_{seq}^{2}}\left(s_{ij}-\sum_{l=1}^{\mathrm{N}}h_{il}h_{jl}\beta_{l}\right)^{2}\right)\label{eq:additive}\end{equation}

\begin{figure}[!tpb]
\centerline{\includegraphics[width=0.5\textwidth]{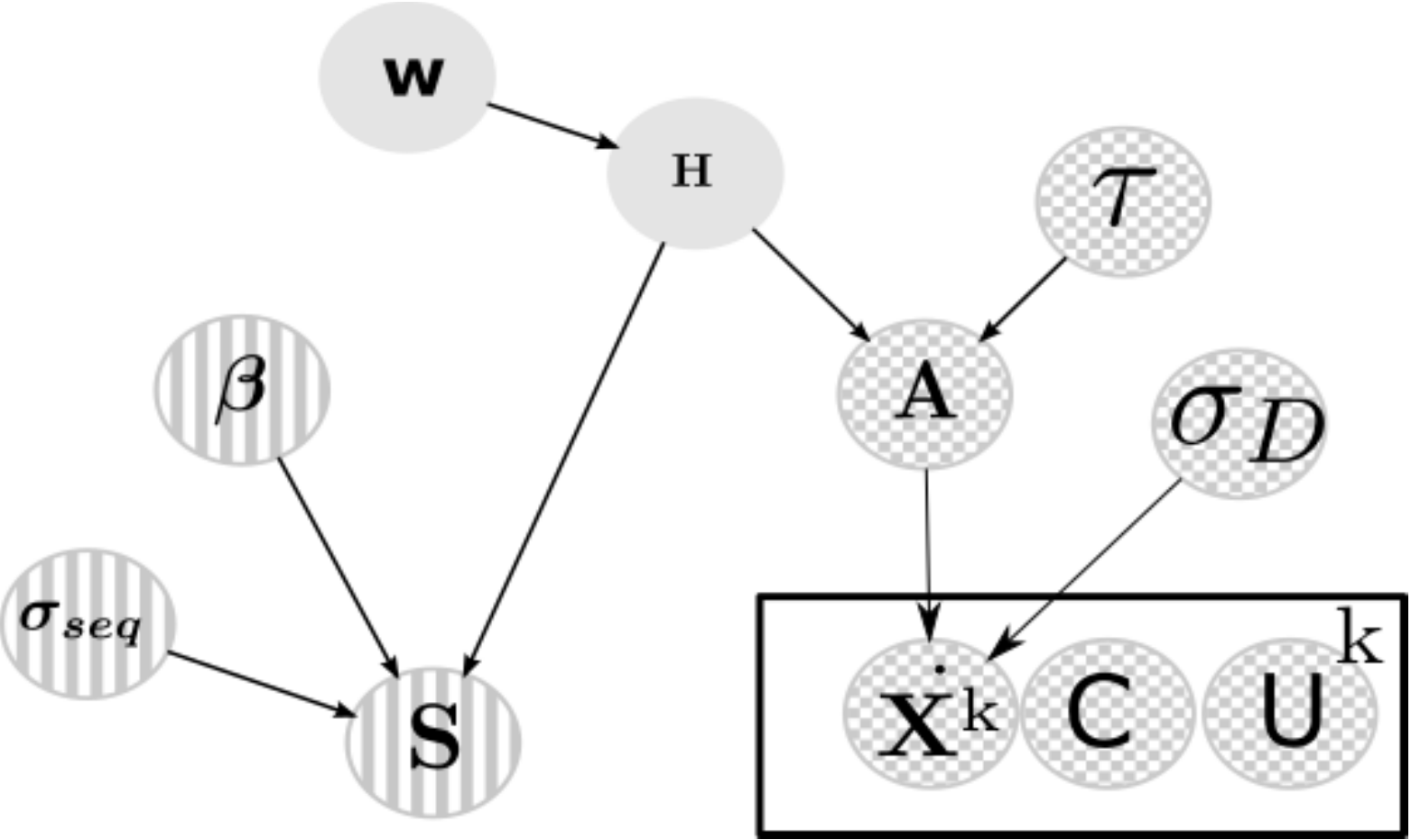}}
\caption{Hierarchical Bayesian model, on top of the hierarchy (green) lies the 
adjacency matrix $\mathbf{H}$ and sparsity parameter $w$. In chequered circles the frequency-domain 
gene expression model and its parameters. In yellow the stripes sequence similarity and its parameters.}\label{fig:model}
\end{figure}

Here the parameter $\left\{ \beta_{l}\right\} \;1\leq l\leq N$ is the
similarity ``induced'' by the $l-th$ transcription factor (a
proportionality constant), and the product $h_{il}h_{jl}$ equals 1 if and only if genes
$i$ and $j$ are both regulated by $l$. This model is a form of additive
clustering \cite{mirkin_additive_1987}. By conditioning on $\mathbf{H}$, we can derive the distribution $p(\beta_l|...)$, which is a Gaussian with non-negative constraints, (see appendix eq. 4). This distribution can be used for sampling posterior values of $\beta$; in our applications, however, we preferred to fix the value of $\beta$ to its non-negative maximum likelihood solution, effectively approximating this conditional posterior with a $\delta$ function. The similarity score variance $\sigma_{seq}$ is given a weakly informative inverse Gamma prior. By completing the square we can derive a Gaussian distribution for the $beta_{l}$ parameters, for its derivation and estimation see appendix section 1. The overall structure of the model is depicted graphically in
Fig. \ref{fig:model}.

\subsection{Inference}
Inference of parameters $\left\{ \mathbf{A,C},\mathbf{H},\sigma_{D},w,\boldsymbol{\tau}\right\} $
is done through a simple Gibbs sampling scheme. Given conjugacy among distributions,
sampling of these parameters is straightforward for all distributions except $\mathrm{p}\left(\beta_{l}\right)$. This distribution is not conjugate, so a Metropolis within Gibbs would be necessary for exact inference. In order to improve performance and given the fact that retrieving the distribution over  $\beta_{l}$ is not an objective; we use the non-negative least square estimate for the vector $\boldsymbol{\beta}$. Convergence was tested by applying Geweke diagnostic over the last 1000 samples of matrix $\mathbf{H}$.  Mathematical derivations of the required conditional posteriors and the general sampling algorithm are described in the appendix.

\section{Results}

In this section we assess the performance of our method on two realistic simulated data sets and a real data set, comparing its performance to two other state of the art methods. We call our method DSS, for DFT-based Spike
and Slab model. The first simulated data set was generated from a well known model
for the \emph{A. thaliana} circadian
clock network \cite{pokhilko_data_2010}. This model is a non-linear ODE-based model which exhibits regular oscillations (for suitable parametrisations), thus matching one of our main modelling assumptions. However, it is a non-linear model, hence introducing an element of model mismatch.
As a second synthetic benchmark data set we used one of the data sets provided by the DREAM 4 challenge
\cite{marbach_2010}. This is again a non-linear model, which exhibits damped oscillatory dynamics in some of the nodes; thus, this data set presents considerably more elements of model mismatch. The last experiment tested the method on a 
real data set of gene expression levels obtained
in a micro-array experiment for the \emph{S. cerevisae} cell cycle transcriptional network \cite{orlando_global_2008}.

Results were assessed in terms of area under the Precision-Recall (AUPR) curve; PR curves plot the fraction of correctly called instances versus the ratio of true positives over true positives plus false negatives. An ideal classifier would give a AUPR of 1, while a random baseline would return the ratio of positives negatives. Inference of the models parameters was conducted by Gibbs Sampling
from the model presented in (Fig. \ref{fig:model}) . In total, 5000 samples
were obtained. The last 1000 samples were selected and averaged to
compute the conditional probability of a link $\mathrm{p}\left(h_{ij}=1|\cdot\right)$
given the model and the expression data, see appendix sections 1.1.1 and 1.1.2 for details into the inputs and outputs of the program. 

\subsection{Competing methods}
As a first comparison, in order to establish the validity of our claim that frequency domain analysis is beneficial for oscillatory networks, we sought to compare our results with a complete analogue in time domain. To do this, we implemented a spline-based alternative to the DFT, using cubic splines interpolation as means of computing the time domain derivative, while the rest of the hierarchical model was left unchanged.
As competing methods  to assess the performance of DSS we selected  
GENIE3 \cite{huynh-thu_inferring_2010}, which is based on random forests, and the ODE-regression
based Inferelator \cite{bonneau_inferelator:_2006,greenfield_robust_2013}.

In a network of  N genes, GENIE3 solves N regression problems by predicting, using random forests, the  
expression level  of each gene as a function of the other N-1 genes (putative regulators). Then  the relative importance of each gene expression 
is evaluated and the putative gene interactions are  ranked.
GENIE3 was designed for
steady state data,  but time-series adaptation can be readily derived and was provided to us by one of the authors. 

The Inferelator  estimates
the parameters of an ODE system using regression with L1-regularization over a finite element approximation of the derivative.
The method has been extended \cite{greenfield_robust_2013}, with new functionalities to incorporate prior information over the network links, and to use alternative optimisation methods for model selection,
including the elastic-net (regularization over L1 and L2 norms) and  Bayesian regression with best subset selection.

Finally, as a simple baselines, we implemented a L1 regularised version of the regression problem in equation \eqref{eq:product}, using the LASSO implementation \cite{tibshirani_regression_1994}.

\subsection{\emph{A. thaliana} circadian clock}
\begin{figure}[!tpb]
\centerline{\includegraphics[width=1.0\textwidth, height=0.25\textheight]{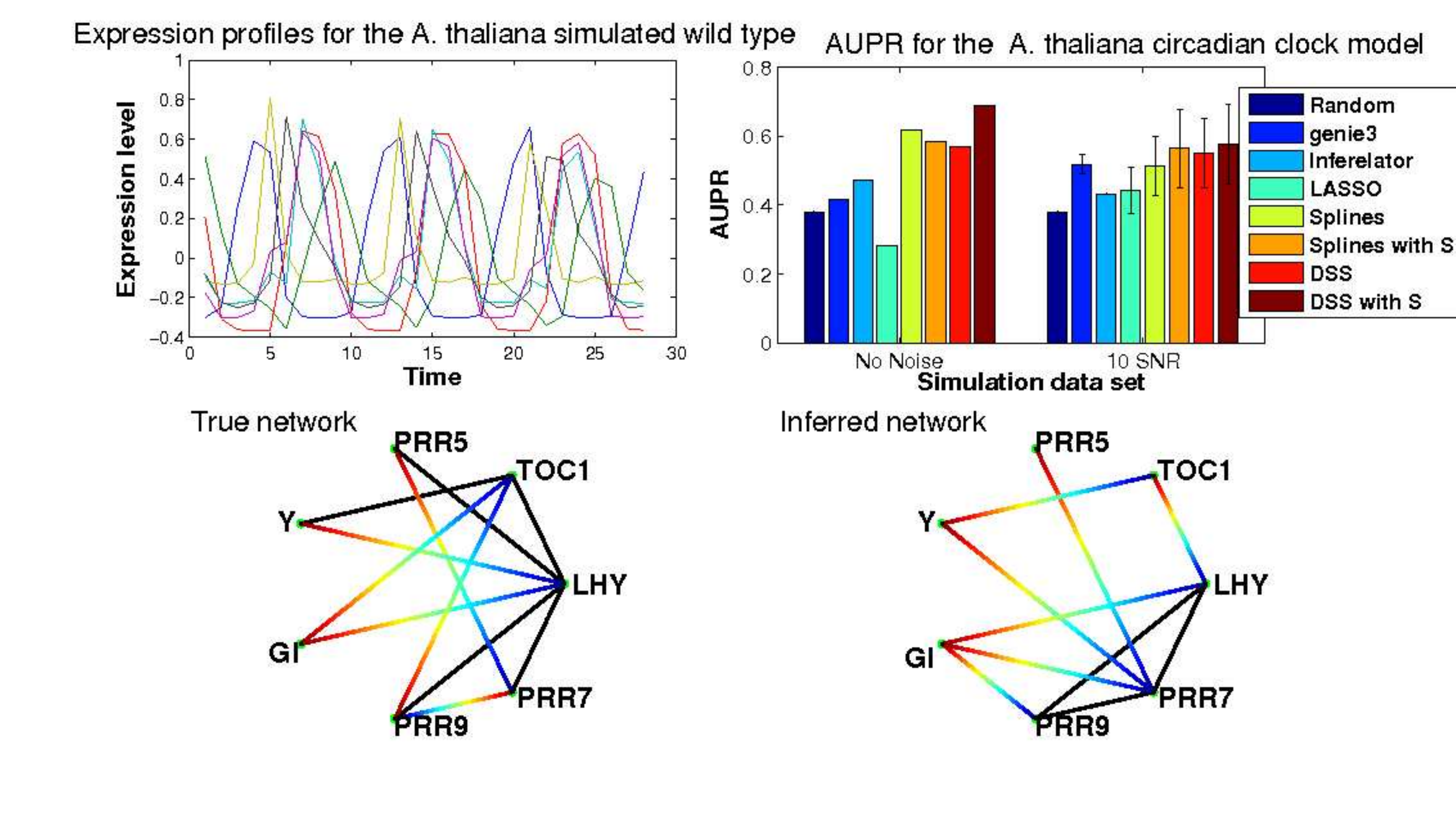}}
\caption{Top left are the simulated gene expression profiles for the wild type data set with SNR 100. Top right are the AUPR values for the 2 different noise levels. Bottom left is the true network topology, going from blue (regulators) to red (targets). Bottom right is the inferred network topology obtained by setting a threshold of 0.5 over the inferred matrix $\mathbf{H}$}\label{fig:circ}
\end{figure}

As a first example we used data generated from a well known oscillatory network model,
the \emph{A. thaliana} circadian clock. The data consists
of simulated mRNA measurements from the model found in \cite{pokhilko_data_2010}.
This non-linear model has 7 transcription factors and 2 post transcriptional elements ZTL and LHYmod. In order to replicate experimental conditions, we assume that only mRNA data is available,
so protein concentrations for the transcriptional and post-transcriptional elements are assumed unobserved. The transcription factors used for network inference are 'LHY ', 'TOC1 ', 'PRR5 ', hypothetical gene 'Y ', 'GI ', 'PRR9' and 'PRR7', the post-transcriptional elements are not considered. 
A graphical representation of the model can be observed in (appendix Fig. 1).
This model was simulated for 3 cycles obtaining 28 samples. The procedure was performed with a light/dark photo period of 12/12, 6/18, 8/16, 18/6 and 20/4 hours which are represented in our model by binary input signals $\mathbf{U}$. This design of our study is created to mimic a realistic experimental setting as in \cite{Edwards_2010}; the biological rationale for such design is that stimulating the system with these different inputs may tease out the contribution of the main drivers of the clock at different times of day.
We also simulated knock-out mutants $\Delta$TOC1, $\Delta$PRR7PRR9,
$\Delta$LHY and $\Delta$GI by the same procedure as presented
in \cite{pokhilko_data_2010} with photo periods of 12/12 hours. These experiments amount to 14 time series; as these data are directly the outputs of an ODE model (without any additional noise) we define this idealised data set as the {\it noiseless} data set. An  additional dataset of 14 time series was  produced by adding Gaussian white noise with a Signal-to-noise (SNR) of 10. An example of the simulated expression levels is plotted in the upper left panel of Fig \ref{fig:circ}.

Using the model specification as ground truth, we proceeded to draw
the PR-curves for the different methods and computed the area under the PR-curve for all the resulting networks. These areas are plotted for the noiseless (simulated data without added noise) and noisy data in the upper right panel of Fig. \ref{fig:circ}. For the case of noisy data, we generated 20 data sets by adding 10SNR-white-Gaussian noise to the noiseless data set. The simulated data has a very well defined oscillatory behaviour, as such the DSS method proved
superior over the alternatives, as can be appreciated in the chart.
The DSS method achieved an   AUPR of 0.57 and  mean 0.55 with std of 0.1 in the noisy case, being  over the mean GENIE3 and Inferelator  estimates with these later yielding less variance in their scores (0 variance in case of the Inferelator). The LASSO solution to the quadratic form also shows an improvement over both methods, even though DSS still has the best results. Interestingly, the spline solution achieves comparably good performance with DSS for the noiseless case, but  under performs in the noisy case. We speculate that this may be due to the sensitivity of the spline estimate of the derivative to noise, confirming our intuition that adopting a set of basis functions well adapted to the problem may convey an advantage. It is intriguing that the method  actually obtains a better performance on  noisy data sets; we speculate that this may be due to the fact that adding noise alleviated the effects of model mismatch (resulting from the LTI approximation). Intuitively, in the absence of noise the attempts to fit non-linear data with a linear model could become more problematic.

To test the effect of including side information, we  simulated a between-gene similarity matrix by drawing $\beta_{l}$ from a uniform distribution $\mathrm{U(0.1,0.6)}$ and using Equation \ref{eq:additive}. In this case we notice an important improvement
by observing an increment in the AUPR to 0.68 in the noiseless case and a mean of 0.57 with std of 0.1 for the noisy data. The principal objective of using this simple simulated similarity matrix 
was to confirm that structural information can be retrieved and used as aid for inference. By clustering the
 co-regulated elements we added additional structural constraints into the inference scheme.  
 
 Finally we included a graphical representation of the true network (Fig. \ref{fig:circ} bottom left) and a network resulting from setting a threshold of 0.5 over the inferred matrix $\mathbf{H}$ (DSS-with-similarity, noiseless data) (Fig. \ref{fig:circ} bottom right). We notice that the 0.5 threshold, while reasonable, is  still arbitrary and is used here only for the purposes of graphical visualisation. The full output from the method is a probability over the existence of edges, and can be better visualised as a heatmap, see appendix sections 1.2 and 2. Directed edges go from blue (regulators) to red (targets), black edges mean bidirectional regulation. As can be appreciated important features such as the bidirectional regulation between 'LHY'-'PRR7' and 'LHY'-'PRR9'   are recovered. Errors are related to the roles of 'PRR7' and 'PRR9' regulating 'GI' instead of 'TOC1'. This may be due to the method confounding the effects of 'TOC1' over these former elements as being closer to the expression patterns of GI. This difficulty discriminating between the roles of the 'PRR' genes is also expressed by inferring  the spurious bidirectional edge between 'PRR7'-'PRR9'.

\subsection{DREAM Challenge}

\begin{figure}[!tpb]
\centerline{\includegraphics[width=1.0\textwidth, height=0.25\textheight]{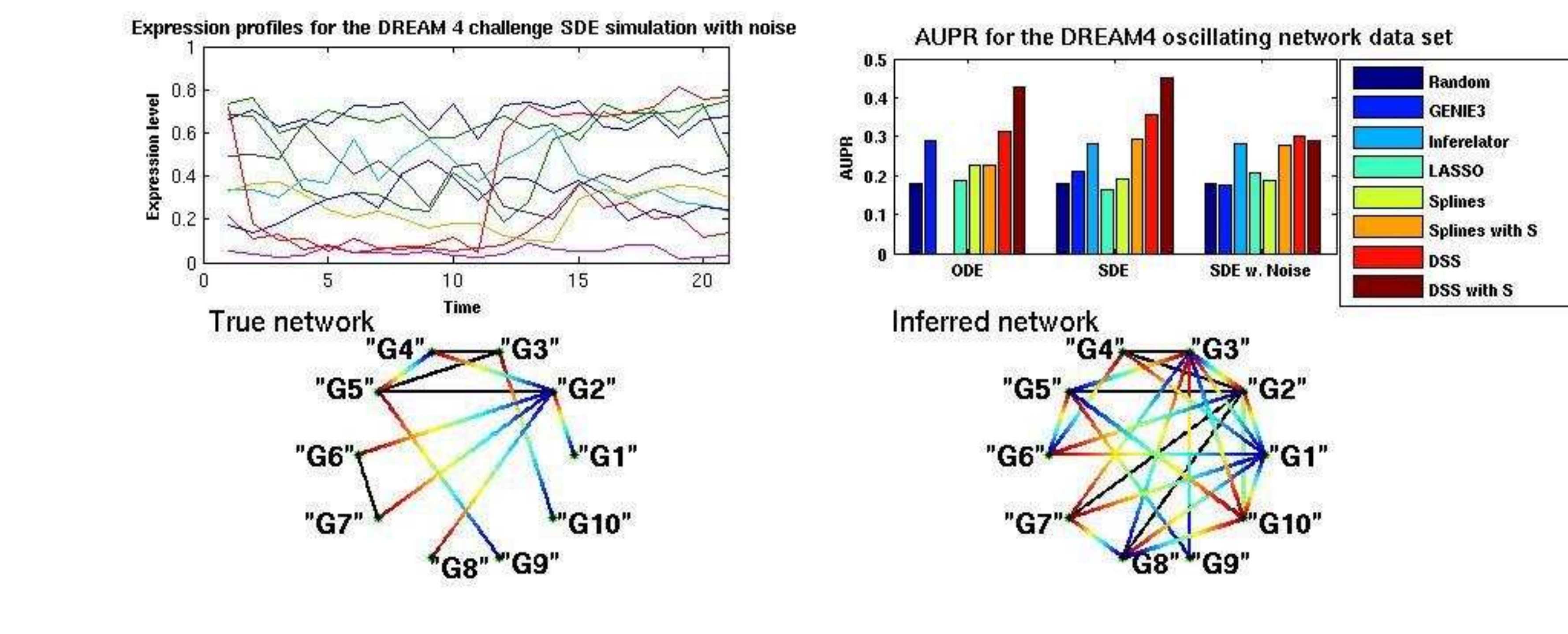}}
\caption{Top left is the expression profiles for the SDE model with experimental noise, node "G9" in red presents a perturbation over half the time points. Top right are the AUPR values for the three simulation models. Bottom left is the true network topology, from blue (regulators) to red (targets). Bottom right is the inferred network obtained by setting a threshold of 0.5 over the inferred matrix $\mathbf{H}$}\label{fig:dream}
\end{figure}

As a second example, we considered a data set from the fourth edition of the DREAM competition \cite{marbach_2010}.This data set is obtained from simulating a 10-node network, of which
three nodes are input nodes; 15 regulatory links are present. Three simulations were present, one with an ODE-based system, another one with a Stochastic Differential Equation (SDE) system and a third one with SDE-based system and added experimental noise. Five time series
are provided for each system, a time series contains 21 samples. The network is subjected to
a single node perturbation, which mathematically corresponds to a change in the basal expression parameter, so the mean expression level of the node changes for half of the time points. The expression profiles for the set of ten genes in one time series is presented in (Fig. \ref{fig:dream} top left). This data set does not comply with the
main assumption of the model (it shows irregular damped oscillations); we therefore expect performance
not to be optimal, but it is still useful to evaluate comparatively the model under such a model mismatch scenario.

Figure (Fig. \ref{fig:dream}) shows a comparison of the area under the P-curve for the three simulated systems. Of these, DSS achieves better performance in the ODE-based simulation, by having an AUPR of 0.31,  higher than the nearest best method (GENIE3). Inferelator could not be executed on this data set due to numerical issues (some expression levels are exactly zero in this example). 
The performance improved for the SDE based simulation, by  achieving an AUPR of 0.35, above inferelator's 0.27. Slightly worse results were achieved for the SDE model with experimental noise, achieving an AUPR of 0.3. By simulating a sequence similarity matrix performance was improved for both ODE and  SDE solutions. In the case of SDE the solution improved dramatically to 0.42. 

As in the previous experiment, the network and its inferred counterpart are presented in Fig \ref{fig:dream} bottom left and bottom right respectively. The inferred network is obtained by setting the threshold to 0.5 over the inferred adjacency matrix for the SDE data with added similarity matrix. As can be observed in the true network, nodes "G1" and "G10" are constant inputs. Node "G9" is subjected to perturbation for half the  time points, thus its effect is propagated through the network by node "G5". 

In the inferred network we can observe some interesting characteristics. First, nodes "G1" and "G9" are identified as input nodes, node "G10" is incorrectly identified as an output only node. Node "G2" maintains its out-degree of 4 even though it's regulators are not correctly identified. Nodes "G9" and "G5" are shown with increased in and out-degree, this may also be due to the confounding effects of the their "parent-son" relationship, specially considering that the perturbed "G9" node has the biggest amplitude of the gene expression profiles, as appreciated by the red curve in the top left plot in Fig. \ref{fig:dream}. 
 
\subsection{\emph{S. cerevisae} cell cycle}
For the last experiment we used a real time series data set collected during the \emph{ S. cerevisiae}
cell cycle. Our evaluation is based on the genes identified
by \cite{haase_topology_2014,orlando_global_2008} and some of their interactions
on the dynamical model found in \cite{chen_integrative_2004}. The
main transcriptional elements selected were 'SWI5', 'YHP1', 'SWI4',
'FKH1', 'SIC1', 'ACE2', 'YOX1', 'STB1', 'NRM1', 'WHI5', 'FKH2', 'MCM1',
'SWI6', 'HCM1', 'NDD1' and 'MBP1'. Their putative regulations were
extracted from literature \{see appendix\} for the putative network used as
ground truth.

The source for the gene expression data is \cite{orlando_global_2008},
it contains 2 wild type replicates and two mutant replicates 
(${\Delta}clb1, 2, 3, 4, 5, 6$) each one containing 14 samples for each gene
 during approximately 2 cell cycles. Additionally, we downloaded promoter sequence
information from \cite{Zhu01071999} for all the network elements. We then proceeded to use the multiple
alignment software Clustal Omega \cite{sievers_fast_2011} to obtain an alignment-based similarity matrix $\mathbf{S}$ between sequences. As an alternative way of encoding sequence information, an alignment-free similarity matrix $\mathbf{S2}$ was built using the method described in \cite{sims_2009}.

\begin{figure}[!tpb]
\centerline{\includegraphics[width=1.0\textwidth, height=0.25\textheight]{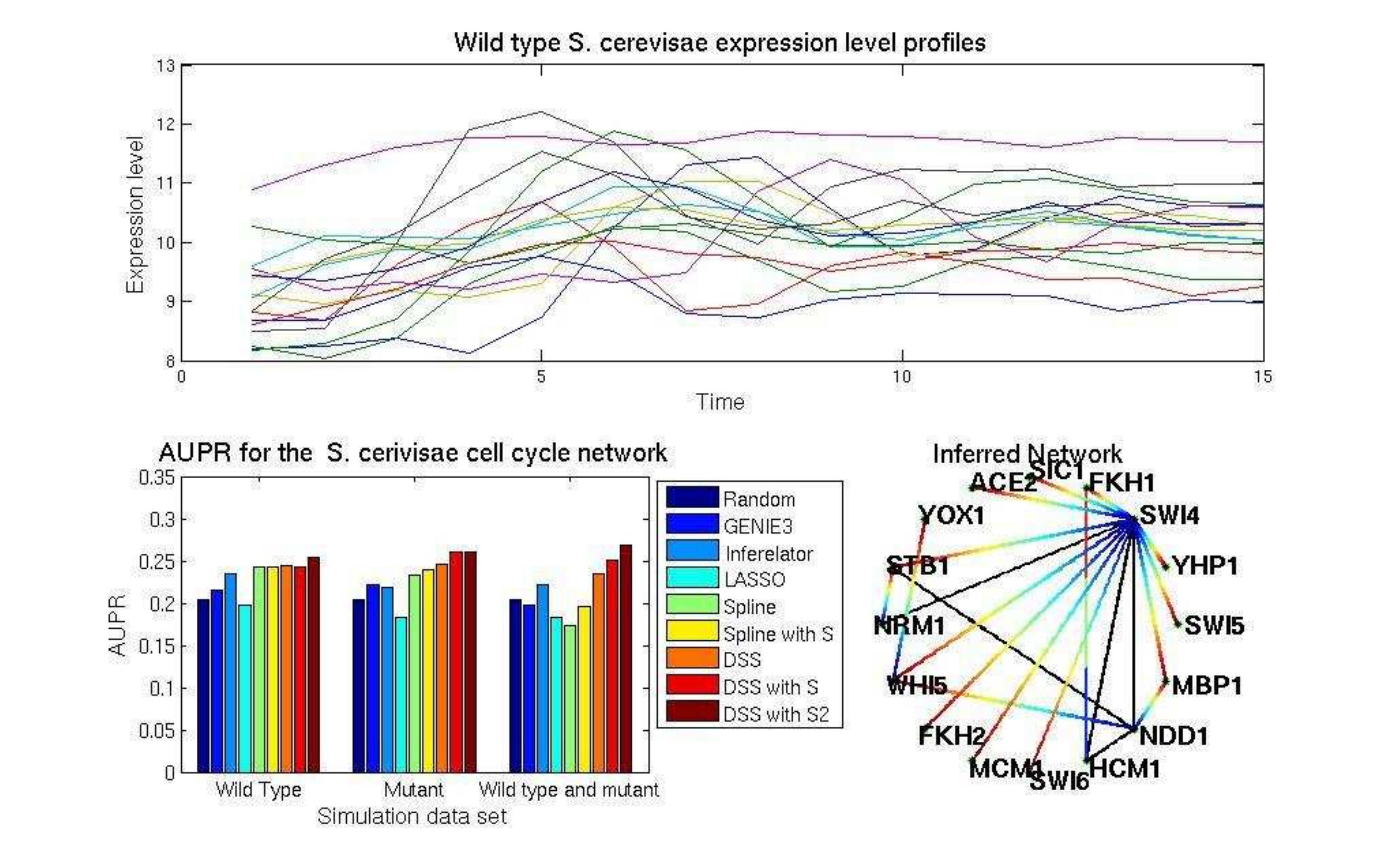}}
\caption{Top wild type yeast expression profiles for the selected genes, bottom left AUPR for the three different data combinations, wild type, mutants, and both. Bottom right network obtained by setting a threshold of 0.9 over matrix $\mathbf{H}$ }\label{fig:yeast}
\end{figure}

We tested three subsets of data, one containing only the wild type expression profiles, other containing only the mutants expression levels, the last data set was the normalized concatenation of both. As an example of the observed gene expression levels, Fig. \ref{fig:yeast} top panel shows the gene expression levels for the wild type conditions.

The AUPR from applying the various methods to this data are presented in Fig. \ref{fig:yeast} bottom left panel. In this case DSS identifies the 
putative network well above the random baseline of 2.1  and above the competing methods. In the case of wildtypes the AUPR of DSS was of 0.24. In the mutant data sets the performance of DSS improves by including sequence similarity achieving an AUPR of 0.2607 and 0.2608 for S and S2 respectively.  The best overall performance was achieved by using the combined data set with sequence similarity matrix S2, resulting in an AUPR of 0.267.

The network in (Fig. \ref{fig:yeast}) bottom right is obtained by setting  the threshold of 0.9 to the inferred network from the combined wild type and mutant dataset with added similarity matrix. In this case FKH1 has a central role in the inferred network, being fully connected to the other elements. Even though this fully connected position is biologically implausible, it does reinforce the important role of FKH1 in the cell cycle, e.g. its role in regulating the M-phase response \cite{kumar_forkhead_2000}.
Another noticeable inferred link concerns the post transcriptional regulation
of SWI6 by WHI5p  \cite{turner_cell_2012}; this regulation was also considered as part of the ground truth network, as in the case of the yeast cell cycle transcriptional and post transcriptional regulations are  intertwined \cite{haase_topology_2014}. Also worth noticing
the regulation of SWI6 by YOX1 (member of the SBF complex) even though evidence suggests causality may be in the opposite direction \cite{venters_comprehensive_2011}.
SWI4 and SWI6 form part of transcription factor complexes SBF and MBF, as such, their regulations may be confounded. This can be appreciated in the regulation of NRM1 by SWI4 in the inferred
network, when in fact NRM1 appears to be regulated by SWI6 \cite{dejesus_hidden_2013}. 
 The transcriptional activator NDD1 is essential during the S-phase \cite{loy_ndd1_1999},  NDD1p along MCM1p bind to FKH2p \cite{haase_topology_2014}, this effect may be observable in the inferred  network by directed edges from NDD1 to YOX1 and from YOX1 to FKH2. 
 
 By observing the AUPR plot we see that mutant data appears to be more informative in this case than wild type, being only marginally inferior to the combined data set with similarity matrix. Part of the experimental design in selecting mutations in \cite{orlando_global_2008} was aiming at attenuating the effects of the post-transcriptional elements of the cell cycle; the stronger performance of our method on the mutant data sets may be explained by this experimental design.
 
Generally, the DSS solution will find the most relevant edges in the network to explain the observed dynamics, while the DSS with similarity method will find the most relevant solution that includes a grouping of the proposed edges according to the similarity matrix. So both results can be analysed separately and may offer additional insight over the whole network behaviour. With this purpose the six inferred networks and the putative ground truth are included in (appendix Fig. 3) for analysis.

\section{Discussion}
Inference of gene regulatory networks from expression data is one of the best studied problems in systems biology. Despite this considerable collective effort, the general problem remains ill-posed and, in the absence of extensive data sets and strong domain expertise, a solution to this problem remains elusive. In this light, it is of interest to consider more delimited problems which may be amenable to specialised but more effective solutions. Oscillatory systems present a prime example of such a problem: while they obviously constitute a specialised subset of regulatory networks, in our opinion they are sufficiently widespread to warrant tailor-made solutions. DSS couples a simplified mechanistic approach (LTI) with frequency domain information to provide such a method.
LTI methods in the time domain for {\it A. thaliana} with experimental data have been studied in \cite{dalchau_understanding_2012}. Our results on the  circadian clock simulation suggest that this frequency domain approach can indeed be fruitful when the model assumptions are reasonably met. As  Results over the DREAM and {\it S. cerevisae} data sets suggest that the method can perform competitively with state of the art methods also when the model assumptions are not precisely met (damped oscillatory behaviour); however, in these cases the methods competitive advantage is smaller or inexistent.  

The use of derivative and ODE information in a network inference framework has some precedents. A method that is in spirit similar to our approach is Inferelator \cite{bonneau_inferelator:_2006}. It casts the network inference problem as
a a parameter inference problem over a first order differential equation
system, then estimates the system parameters via regularized regression
over a finite differences solution to the system. 
Recently Bayesian approaches that make use of the derivative information have also been proposed. In \cite{oates_network_2012} a probabilistic model for integrating a linearised version of network dynamics in a regression framework is presented. \cite{DondelingerAISTATS13} attacked the problem of parameter inference of an ODE system jointly with a Bayesian regression over the gene expression levels. The basis of this model is a Gaussian process with product of experts likelihood, not dissimilar from our model in equation \eqref{eq:product}. However, the authors in \cite{DondelingerAISTATS13} did not attempt a joint parameter estimation and variable selection problem, stopping short of formulating the problem in terms of network inference. Basis functions in time domain (splines) have already been applied to network inference problems in systems biology, primarily to model unknown non-linear transition functions \cite{grenits}. The distinctive part of our work is the proposal of a frequency domain approach for oscillatory systems, and in particular the embedding of our method within a hierarchical framework where integration of additional information is natural. We expect that non-linearities encoded as basis functions as in \cite{grenits} would be a valuable extension of our work and likely result in an improvement in performance.

While we believe that the DSS method provides promising results, there are several inherent limitations in our approach. Importantly, the LTI approximation implies that self regulation
is confounded with decay, so such types of interactions cannot be identified. Empirical results also seem to suggest that post transcriptional interactions may be confounded with transcriptional interactions; this is to be expected, as post-transcriptional interactions are not modelled in our framework. For such reasons, direct application to models that include complex post-transcriptional interactions, such as \cite{pokhilko_clock_2012}, is not advised. Furthermore, as all Bayesian network inference methods, DSS also suffers from multi-modal posterior distributions. The use of auxiliary information, such as sequence similarity, can be beneficial to ameliorate this problem. Many different types of auxiliary information can be considered, and indeed alternative models for incorporating sequence similarity could also be used. A major strength of a Bayesian hierarchical model is that different models for auxiliary information could be easily incorporated within the DSS framework.

\section*{Acknowledgement}
We thank Botond Cseke, V\^an-Anh Huynh-Thu and Daniel Seaton for useful discussions.  The GENIE3 software adapted for time series data was kindly provided to us by Dr V\^an-Anh Huynh-Thu.

DTB is funded by a Microsoft Research Studentship. GS acknowledges support from the European Research Council under grant MLCS30699. SynthSys was founded as a Centre for Integrative Biology by BBSRC/EPSRC award D19621 to AJM and others.

\appendix
\newpage
\title{Appendix for A Bayesian approach for Structure learning
of Oscillating Regulatory Networks}

\maketitle

\section{Gibbs sampler for the Hierarchical Bayesian model.}

Likelihood funcion for the LTI system.

\begin{eqnarray*}
\mathrm{p}\left(\mathbf{Q}_{k}|\sigma\right) & \propto & \sigma_{D}^{-\frac{\mathrm{M}}{2}}\exp\left(-\frac{1}{2\sigma^{-2}}\mathrm{tr}\left(\mathbf{Q}_{k}^{\mathrm{T}}\mathbf{Q}_{k}\right)\right)\\
\prod_{k}\mathrm{p}\left(\mathbf{Q}_{k}|\sigma\right) & \propto & \sigma_{D}^{-\frac{\mathrm{M+\mathrm{K}}}{2}}\exp\left(-\frac{1}{2\sigma^{-2}}\sum_{k}\mathrm{tr}\left(\mathbf{Q}_{k}^{\mathrm{T}}\mathbf{Q}_{k}\right)\right)\\
\mathbf{Q}_{k} & = & \left(\mathbf{\dot{X}}_{k}-\left[\begin{array}{cc}
\mathbf{X}_{k} & \mathbf{U}_{k}\end{array}\right]\left[\begin{array}{c}
\mathbf{A}^{\mathrm{T}}\\
\mathbf{C}^{\mathrm{T}}\end{array}\right]\right)\end{eqnarray*}

Conditional distribution of $\left[\begin{array}{c}
\mathbf{A}^{\mathrm{T}}\\
\mathbf{C}^{\mathrm{T}}\end{array}\right]$

Let's call $\mathbf{B}=$$\left[\begin{array}{c}
\mathbf{A}^{\mathrm{T}}\\
\mathbf{C}^{\mathrm{T}}\end{array}\right]$ and $\mathbf{R}_{k}=\left[\begin{array}{cc}
\mathbf{X}_{k} & \mathbf{U}_{k}\end{array}\right]$

\begin{eqnarray*}
\mathrm{p}\left(\left\{ \mathbf{\dot{X}}_{k}\right\} ,\left\{ \mathbf{R}_{\mathrm{k}}\right\} |\mathbf{B},\sigma_{D}\right) & \propto & \prod_{k}\mathrm{p}\left(\mathbf{\mathbf{\dot{X}}}_{k}|\mathbf{R}_{k}\mathbf{B},\sigma\right)\mathrm{p}\left(\mathbf{B}|\mathbf{H},\boldsymbol{\tau}\right)\\
 & \propto & \prod_{\mathrm{k}}\exp\left(-\frac{1}{2\sigma_{D}^{2}}\mathrm{Tr}\left(-\mbox{2}\left(\mathbf{R}_{k}^{\mathrm{T}}\dot{\mathbf{X}}_{k}\right)^{\mathrm{T}}\mathbf{B}+\mathbf{B}^{\mathrm{T}}\left(\mathbf{\mathbf{R}}_{k}^{\mathrm{T}}\mathbf{R}_{k}\right)\mathbf{B}\right)\right)\\
 & \propto & \exp\left(-\frac{1}{2\sigma_{D}^{2}}\mathrm{Tr}\left(-2\boldsymbol{\eta}^{\mathrm{T}}\mathbf{B}+\mathbf{B}^{\mathrm{T}}\boldsymbol{\Psi}\mathbf{B}\right)\right)\\
\boldsymbol{\eta}^{\mathrm{T}} & = & \left(\sum_{\mathrm{k}}\mathbf{R}_{k}^{\mathrm{T}}\dot{\mathbf{X}}_{k}\right)\\
\boldsymbol{\Psi} & = & \left(\sum_{\mathrm{k}}\mathbf{\mathbf{R}}_{k}^{\mathrm{T}}\mathbf{R}_{k}\right)\end{eqnarray*}
 Now by using the vectorization transformation for an arbitrary matrix
$\mathbf{M}$, such that $\bar{\mathbf{M}}=\mathrm{vec}\left(\mathbf{M}\right)$
and properties

\begin{eqnarray*}
\mathrm{Tr}\left(\mathbf{M}^{\mathrm{T}}\mathbf{N}\right) & = & \bar{\mathbf{M}}^{\mathrm{T}}\bar{\mathbf{N}}\\
\mathrm{vec}\left(\mathbf{M}\mathbf{N}\right) & = & \left(\mathbf{I}\otimes\mathbf{M}\right)\bar{\mathbf{N}}\end{eqnarray*}

Then we have

\begin{eqnarray*}
\mathrm{p}\left(\left\{ \mathbf{\dot{X}}_{k}\right\} ,\left\{ \mathbf{R}_{\mathrm{k}}\right\} |\mathbf{B},\sigma\right) & \propto & \exp\left(-\frac{1}{2\sigma_{D}^{2}}\left(-2\mathrm{Tr}\left(\boldsymbol{\eta}^{\mathrm{T}}\mathbf{B}\right)+\mathrm{Tr}\left(\mathbf{B}^{\mathrm{T}}\boldsymbol{\Psi}\mathbf{B}\right)\right)\right)\\
 & \propto & \exp\left(-\frac{1}{2\sigma_{D}^{2}}\left(-2\bar{\boldsymbol{\eta}}^{\mathrm{T}}\bar{\mathbf{B}}+\bar{\mathbf{B}}^{\mathrm{T}}\mathrm{vec}\left(\boldsymbol{\Psi}\mathbf{B}\right)\right)\right)\\
 & \propto & \exp\left(-\frac{1}{2\sigma_{D}^{2}}\left(-2\bar{\boldsymbol{\eta}}^{\mathrm{T}}\bar{\mathbf{B}}+\bar{\mathbf{B}}^{\mathrm{T}}\left(\mathbf{I}\otimes\boldsymbol{\Psi}\right)\bar{\mathbf{B}}\right)\right)\end{eqnarray*}

then by using vectorization we write the prior in canonical form

\begin{eqnarray*}
\mathrm{p}\left(\mathbf{B}|\mathbf{H},\boldsymbol{\tau}\right) & \propto & \exp\left(-\frac{1}{2}\bar{\mathbf{B}}^{\mathrm{T}}\boldsymbol{\Gamma}\bar{\mathbf{B}}\right)\\
\boldsymbol{\Gamma} & = & \mathrm{diag}\left(h_{ij}\tau_{ij}^{2}\right)\end{eqnarray*}

finally multiplying by the gaussian with hypervariance given by the
spike and slab prior and completing the square we get

\begin{eqnarray}
\mathrm{p}\left(\left\{ \mathbf{\dot{X}}_{k}\right\} ,\left\{ \mathbf{R}_{\mathrm{k}}\right\} |\mathbf{B},\sigma\right)\mathrm{p}\left(\mathbf{B}|\mathbf{H},\boldsymbol{\tau}\right) & \propto & \exp\left(-\frac{1}{2\sigma^{2}}\left(-2\bar{\boldsymbol{\eta}}^{\mathrm{T}}\bar{\mathbf{B}}+\bar{\mathbf{B}}^{\mathrm{T}}\left(\mathbf{I}\otimes\boldsymbol{\Psi}\right)\bar{\mathbf{B}}+\bar{\mathbf{B}}^{\mathrm{T}}\sigma^{2}\boldsymbol{\Gamma}\bar{\mathbf{B}}\right)\right)\nonumber \\
\mathrm{p}\left(\mathbf{B}|\left\{ \mathbf{\dot{X}}_{k}\right\} ,\left\{ \mathbf{R}_{\mathrm{k}}\right\} ,\mathbf{H},\boldsymbol{\tau},\sigma\right) & \sim & \mathcal{N}\left(\bar{\boldsymbol{\mu}},\sigma^{2}\boldsymbol{\Sigma}\right)\label{eq:bcond}\\
\bar{\boldsymbol{\mu}} & = & \boldsymbol{\Sigma}^{-1}\bar{\boldsymbol{\eta}}^{\mathrm{T}}\label{eq:mu}\\
\boldsymbol{\Sigma^{-1}} & = & \mathbf{I}\otimes\boldsymbol{\Psi}+\sigma^{2}\boldsymbol{\Gamma}\label{eq:sigma}\end{eqnarray}

Spike and slab prior over the coefficients

\begin{eqnarray}
\mathrm{p}\left(b_{ij}|h_{ij},\tau\right) & \sim & \mathcal{\mathcal{N}}\left(0,h_{ij}\tau_{ij}^{2}\right)\label{eq:prior}\\
\mathrm{p}\left(h_{ij}\mathrm{|}w\right) & \sim & \left(1-w\right)\delta_{v0}+w\delta_{1}\nonumber \end{eqnarray}

Conditional distribution of $\tau{}^{-2}$

\begin{eqnarray}
\mathrm{p}\left(\tau_{ij}^{-2}|a_{ij},h_{ij}\right) & \sim & \mathrm{Gamma}\left(a_{1}+\frac{1}{2},a_{2}+\frac{b_{ij}^{2}}{2h_{ij}}\right)\label{eq:ptau}\end{eqnarray}

Conditional distribution of $w$

\begin{equation}
\mathrm{p}\left(w|\mathbf{H}\right)\sim\mathrm{Beta}\left(c_{1}+\#\{h_{ij}=1\},c_{2}+\#\{h_{ij}=v_{0}\}\right)\label{eq:tw}\end{equation}

Condtional distribution of $\sigma_{D}^{-2}$

\begin{equation}
\mathrm{p}\left(\sigma_{D}^{-2}|\left\{ \mathbf{Q}_{k}\right\} \right)\sim\mathrm{Gamma}\left(b_{1}+\frac{\mathrm{M}\mathrm{K}}{2},b_{2}+\frac{\sum_{k}\mathrm{tr}\left(\mathbf{Q}_{k}^{\mathrm{T}}\mathbf{Q}_{k}\right)}{2}\right)\label{eq:sigmad}\end{equation}

Likelihood of the similarity scores

\begin{eqnarray}
\mathrm{p}\left(s_{\mathrm{ij}}|\mathbf{H},\boldsymbol{\beta},\sigma_{seq}\right) & \propto & \sigma_{seq}^{-1/2}\exp\left(-\frac{1}{2\sigma_{seq}^{2}}\left(s_{ij}-\sum_{l}^{\mathrm{N}}h_{il}h_{jl}\beta_{l}\right)^{2}\right)\label{eq:s}\end{eqnarray}

having vector$\bar{\mathbf{S}}=\left[s_{ij}\right]_{i<j}$ representing
the off-diagonal elements of the upper triangular matrix of $\mathbf{S}$,
vector$\mathbf{\bar{h}}_{i}$ representing the $i-th$ row vector
of $\mathbf{H}$ and $\circ$ representing the elementwise product
(Hadamard product). Then the distribution of the upper diagonal elements
is:

\begin{eqnarray*}
\mathrm{p}\left(\bar{\mathbf{S}}|\mathbf{H},\boldsymbol{\beta},\sigma_{seq}\right) & \propto & \sigma_{seq}^{-\mathrm{D}/2}\exp\left(-\frac{1}{2\sigma_{seq}^{2}}\left(\bar{\mathbf{S}}-\left[\mathbf{\bar{h}}_{i}\circ\mathbf{\bar{h}}_{j}\right]_{i<j}\boldsymbol{\beta}\right)^{\mathrm{T}}\left(\bar{\mathbf{S}}-\left[\mathbf{\bar{h}}_{i}\circ\mathbf{\bar{h}}_{j}\right]_{i<j}\boldsymbol{\beta}\right)\right)\end{eqnarray*}

Conditional distribution for $\sigma_{seq}^{2}$

\begin{eqnarray}
\mathrm{p}\left(\sigma_{seq}^{-2}|\bar{\mathbf{S}},\boldsymbol{\beta}\right) & \sim & \mathrm{p}\left(\bar{\mathbf{S}}|\mathbf{H},\boldsymbol{\beta},\sigma_{seq}^{-2}\right)\mathrm{p}\left(\sigma_{seq}^{-2}\right)\label{eq:psigma}\\
 & \sim & \mathrm{Gamma}\left(a_{1}+\frac{\mathrm{D^{2}}}{2},a2+\frac{\left(\bar{\mathbf{S}}-\left[\mathbf{\bar{h}}_{i}\circ\mathbf{\bar{h}}_{j}\right]_{i<j}\boldsymbol{\beta}\right)^{\mathrm{T}}\left(\bar{\mathbf{S}}-\left[\mathbf{\bar{h}}_{i}\circ\mathbf{\bar{h}}_{j}\right]_{i<j}\boldsymbol{\beta}\right)}{2}\right)\nonumber \end{eqnarray}

Sampling $\boldsymbol{\beta}$

Instead of sampling from $\mathrm{p}\left(\beta_{l}|\centerdot\right)$
we use the non-negative least square estimate of $\boldsymbol{\beta}$,
by solving the quadratic form

\begin{equation}
\min_{\boldsymbol{\beta}}\left(\bar{\mathbf{S}}-\left[\mathbf{\bar{h}}_{i}\circ\mathbf{\bar{h}}_{j}\right]_{i<j}\boldsymbol{\beta}\right)^{\mathrm{T}}\left(\bar{\mathbf{S}}-\left[\mathbf{\bar{h}}_{i}\circ\mathbf{\bar{h}}_{j}\right]_{i<j}\boldsymbol{\beta}\right);\; s.t.\beta_{l}\geq0\label{eq:betas}\end{equation}

Conditional distribution for $h_{ij}$

We define matrices$\mathbf{H}_{ij}^{v0}=\left[\mathbf{\bar{h}}_{i}\circ\mathbf{\bar{h}}_{j}\right]_{i<j}$
such that $h_{ij}=v0$, and $\mathbf{H}_{ij}^{1}=\left[\mathbf{\bar{h}}_{i}\circ\mathbf{\bar{h}}_{j}\right]_{i<j}$
such that $h_{ij}=1$.

\begin{eqnarray*}
\mathrm{p}\left(h_{ij}|h_{/ij},\boldsymbol{\beta},\sigma_{seq}^{-2},\mathbf{S},\mathbf{B},w,\tau,v_{0}\right) & = & P\left(\mathrm{h}_{ij}\mid w\right)P\left(a_{ij}\mid h_{ij},\tau_{ij}^{2}\right)\mathrm{p}\left(\bar{\mathbf{S}}|h_{/ij},h_{ij},\boldsymbol{\beta},\sigma_{seq}\right)\end{eqnarray*}

\begin{eqnarray}
\mathrm{p}\left(h_{ij}|\centerdot\right) & \sim & \frac{m_{_{v0}}}{m_{_{v0}}+m_{1}}\left(1-w\right)\delta_{v0}+\frac{m_{1}}{m_{_{v0}}+m_{1}}w\delta_{1}\label{eq:ph}\\
m_{v0} & = & \sigma_{seq}^{-\mathrm{D}/2}v_{0}^{-1/2}\nonumber \\
 & \times & \exp\left(-\frac{1}{2}\left(\frac{b_{ij}^{2}}{v_{0}\tau_{ij}^{2}}+\frac{1}{\sigma_{seq}^{2}}\left(\bar{\mathbf{S}}-\mathbf{H}_{ij}^{v0}\boldsymbol{\beta}\right)^{\mathrm{T}}\left(\bar{\mathbf{S}}-\mathbf{H}_{ij}^{v0}\boldsymbol{\beta}\right)\right)\right)\nonumber \\
m_{1} & = & \sigma_{seq}^{-\mathrm{D}/2}\nonumber \\
 & \times & \exp\left(-\frac{1}{2}\left(\frac{b_{ij}^{2}}{\tau_{ij}^{2}}+\frac{1}{\sigma_{seq}^{2}}\left(\bar{\mathbf{S}}-\mathbf{H}_{ij}^{1}\boldsymbol{\beta}\right)^{\mathrm{T}}\left(\bar{\mathbf{S}}-\mathbf{H}_{ij}^{1}\boldsymbol{\beta}\right)\right)\right)\nonumber \end{eqnarray}

\subsection{Algorithm}

The Algorithm \ref{alg:Algorithm-for-the} presents the Gibbs sampler
for parameter inference in the Hierarchical model. A brief overview
of the inputs and outputs is also presented.

\subsubsection{Program Inputs}

\begin{figure}
\begin{centering}
\includegraphics[width=1\textwidth]{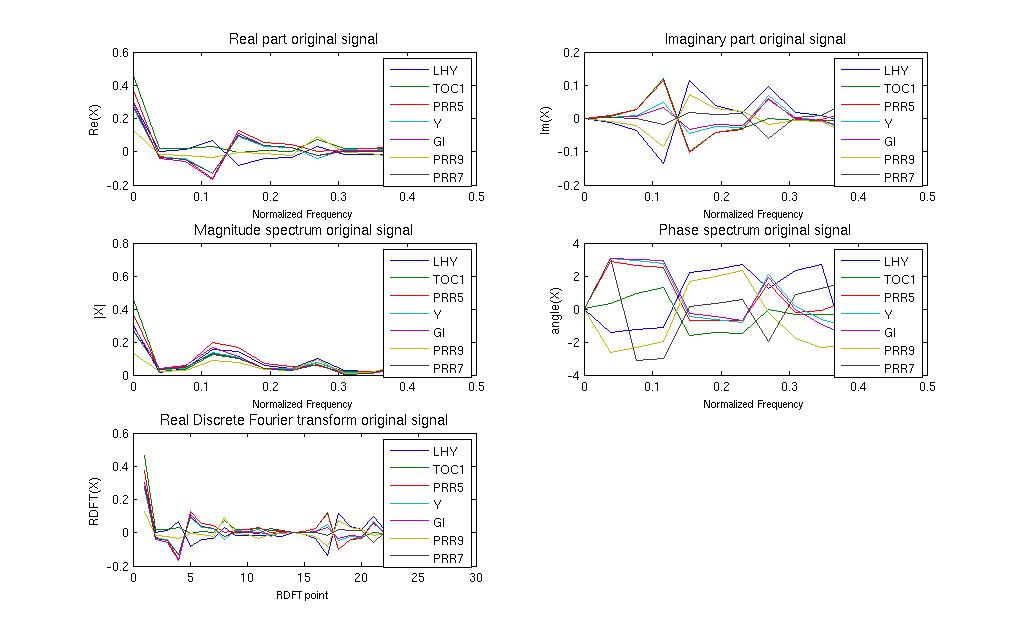}\caption{Spectra for the \emph{A. thaliana }circadian clock simulation with
a 12 hour photo-period\label{fig:Spectra-for-the}}
 
\par\end{centering}

\end{figure}

\begin{figure}
\centering{}\includegraphics[width=1\textwidth]{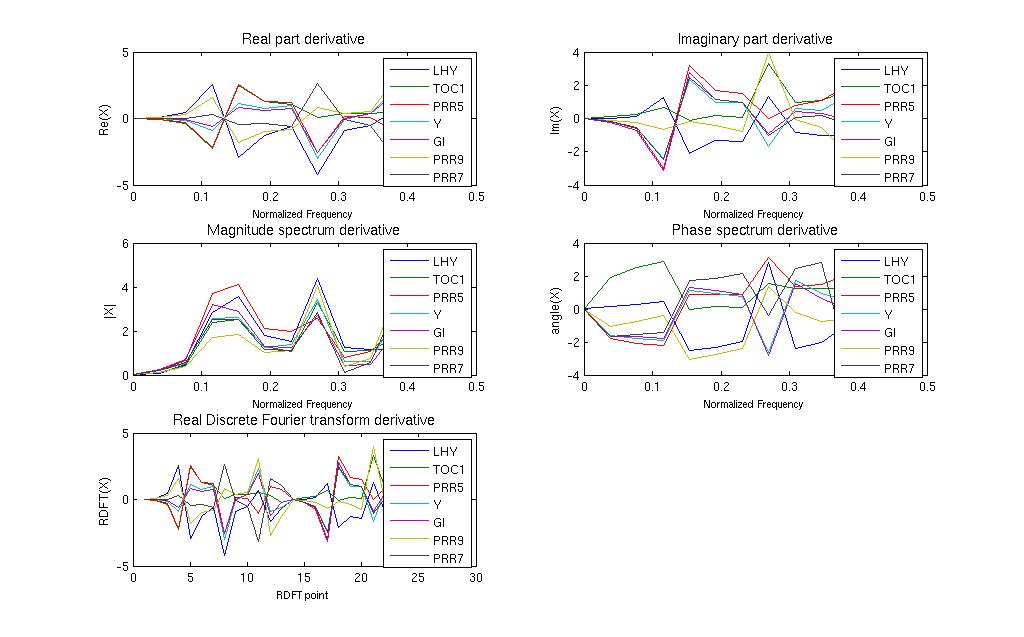}\caption{Derivative Spectra for the \emph{A. thaliana }circadian clock simulation
with a 12 hour photo-period\label{fig:DSpectra-for-the}}
\end{figure}

Due that the DFT computed by the FFT algorithm yields complex numbers,
a real representation of the coefficients is needed. We work with
the so-called Real Discrete Fourier Transform. It consists of stacking
the real over the imaginary part of the first $\mbox{\ensuremath{\left(M-1\right)}/2}$
FFT coefficients. We denote this $M\times N$ matrix $\mathbf{X}$.
Figures (\ref{fig:Spectra-for-the}) and (\ref{fig:DSpectra-for-the})
show plots for the DFT real and imaginary parts of the \emph{A. thaliana
}data-set, with a photoperiod of 12 hours. The magnitude and phase
spectra is also shown, finally the RDFT is presented at the bottom
of the picture.

Hyperparameters $a_{1}$ and $a_{2}$ are set to 1 so $w$ is uniformly
distributed in the interval $\left[0,1\right]$. Parameters $b_{1}$
and $b_{2}$ are set so the hypervariance $h_{ij}\tau^{2}$ has a
continuos bi-modal distribution, according to the recommendations
of \cite{ishwaran}, in which they set them to 5 and 50 respectively.
Alternative parametrization of 50 and 500 was also tested yielding
better results in some cases.

The hyperparameters $c_{1}$ and $c_{2}$for$\sigma_{D}^{-2}$ are
set to 0.001 and 0.001, this is a weak prior reflecting uncertainty
about the linearity of the system. Hyperparamters $d_{1}$ and $d_{2}$
are set to 10 and 0.001, this parametrization required a manual tunning,
as the scale parameter $\sigma_{seq}^{-2}$ having a weak prior resulted
in the effects of the sequence similarity model to banish. By modifying
this prior we can give more {}``weight'' to the sequence similarity
clustering thus, the flexibility of the Hierarchical model.

\subsubsection{\emph{A. thaliana }light inputs U}

\begin{figure}[H]

\includegraphics[width=1\textwidth]{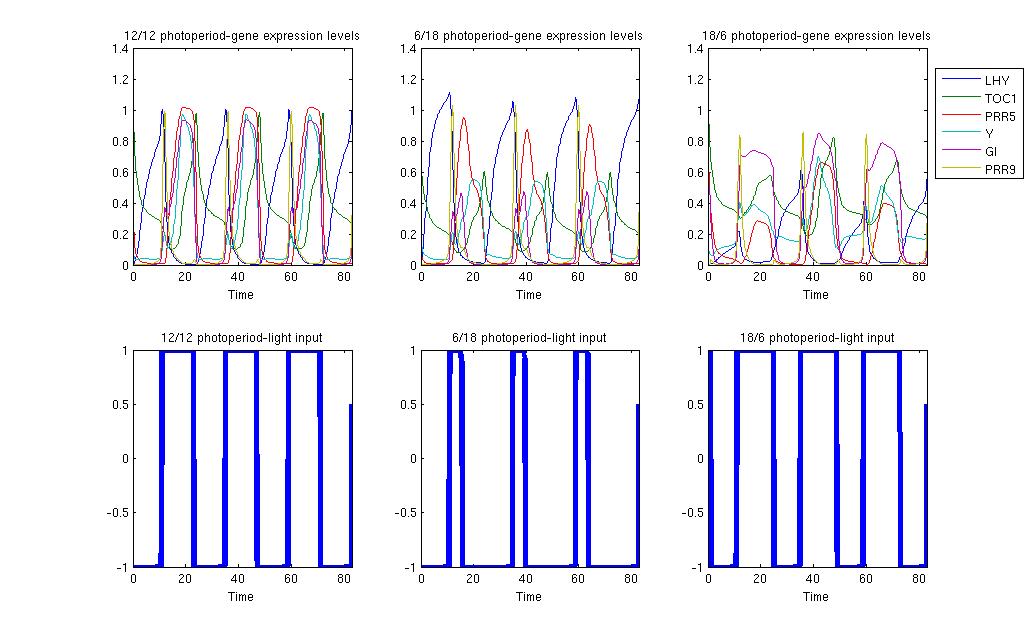}\caption{Three examples of light input for the \emph{A. thaliana} circadian
clock simulation.}

\end{figure}

\subsubsection{Output}

The Gibbs sampler presented in the previous section allows us to draw
samples from the joint conditional distribution $p\left(\mathbf{H},\mathbf{A},\mathbf{C},\beta,w,\tau,\sigma_{D},\sigma_{s}|\left\{ \mathbf{\mathbf{X}_{k}}\right\} \right)$.
The marginal distribution for each of the models parameters can be
drawn from this joint distribution, and the expected value for each
parameter equals to the average of the samples. 

For example, figure illustrates the expected value for matrix $\mathbf{H}$
obtained from averaging over 1000 samples drawn from the marginal
distribution $p\left(\mathbf{H}|\left\{ \mathbf{\mathbf{X}_{k}}\right\} \right)$.
This figure shows in dark red those elements with higher probability
of a regulatory interaction under the model assumptions, except the
diagonal elements, which represent the decay rates of the equation
model. The AUPR were computed by thresholding the off-diagonal elements
of this matrix for each data-set.

\begin{algorithm}
Inputs: $K$ time series of $M$ time-points for $N$ gene expression
levels, encoded in matrices $\left\{ x_{k}\right\} $. Prior hyper-parameters
$a_{1},a_{2},b_{1},b_{2},c_{1},c_{2}$. Optional similarity matrix
$\mathbf{S}$.

Outputs: Joint conditional posterior distribution $p\left(\mathbf{H},\mathbf{A},\mathbf{C},\beta,w,\tau,\sigma_{D},\sigma_{s}|\left\{ \mathbf{\mathbf{X}_{k}}\right\} \right)$
\begin{enumerate}
\item Obtain the DFT of $\left\{ x_{k}\right\} $ and the corresponding
RDFT coefficient matrices $\left\{ \mathbf{\mathbf{X}_{k}}\right\} $
\item Compute the derivatives $\left\{ \mathbf{\dot{\mathbf{X}_{k}}}\right\} $
\item Sample from the conditional distribution over the LTI coefficients,
given in eq. (\ref{eq:bcond})
\item Sample from the conditional distribution over$\tau^{-2}$ given by
eq. (\ref{eq:ptau})
\item Sample $\mathbf{H}$ from eq. (\ref{eq:ph}), to account for the decay
rates we set diagonal elements $h_{ii}$ to 1, and set the diagonal
elements of matrix $\mathbf{A}$ to negative.
\item Sample $w$ from eq. (\ref{eq:tw})
\item Sample $\sigma_{D}$ from eq. (\ref{eq:sigmad})
\item OPTIONAL sample $\sigma_{seq}$ from eq. (\ref{eq:psigma}) and $\beta$
from the nonnegative least squares solution to equation (\ref{eq:betas}).
\item Return to step 3
\end{enumerate}
Note: A burn in period of 4000 samples is considered in the general
purpose implementation of the model.

\caption{Algorithm for the DFT-based Spike and slab prior model with sequence
similarity.\label{alg:Algorithm-for-the}}

\end{algorithm}

\begin{figure}
\centering{}\includegraphics[width=0.6\textwidth]{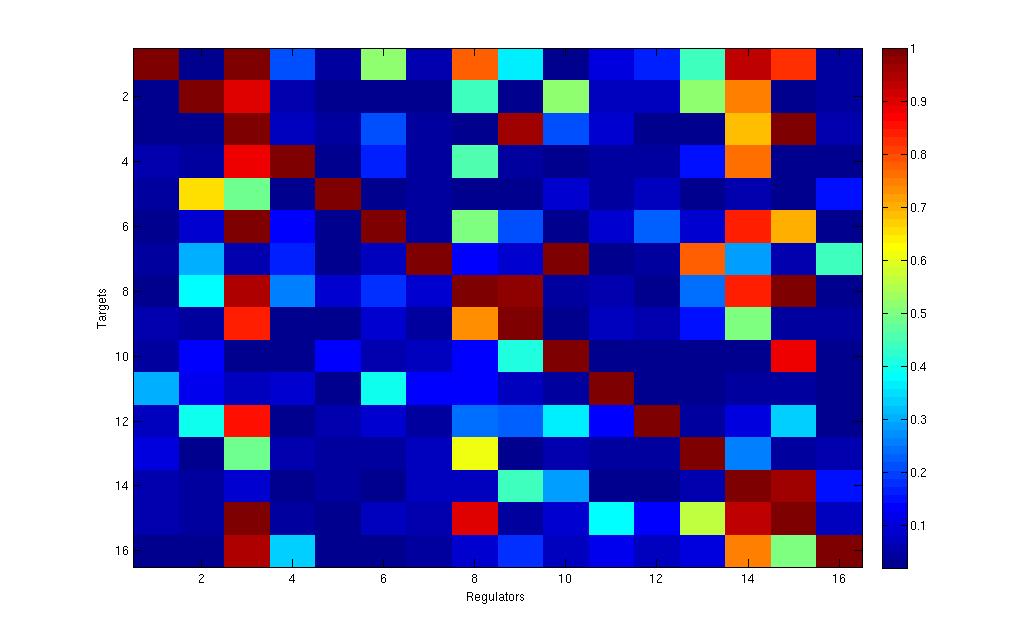}\caption{Heat-map representing the expected value for $p\left(\mathbf{H}|\centerdot\right)$
obtained by averaging the last 1000 samples. Rows represent targets
and columns regulators. The diagonal indicates the decay parameters
$\mbox{\ensuremath{\lambda}}$.\label{fig:Heat-map-representing-the}}
\end{figure}

\section{Heatmaps and PR curves for the experiments}

\begin{figure}[H]
\includegraphics[width=1\textwidth]{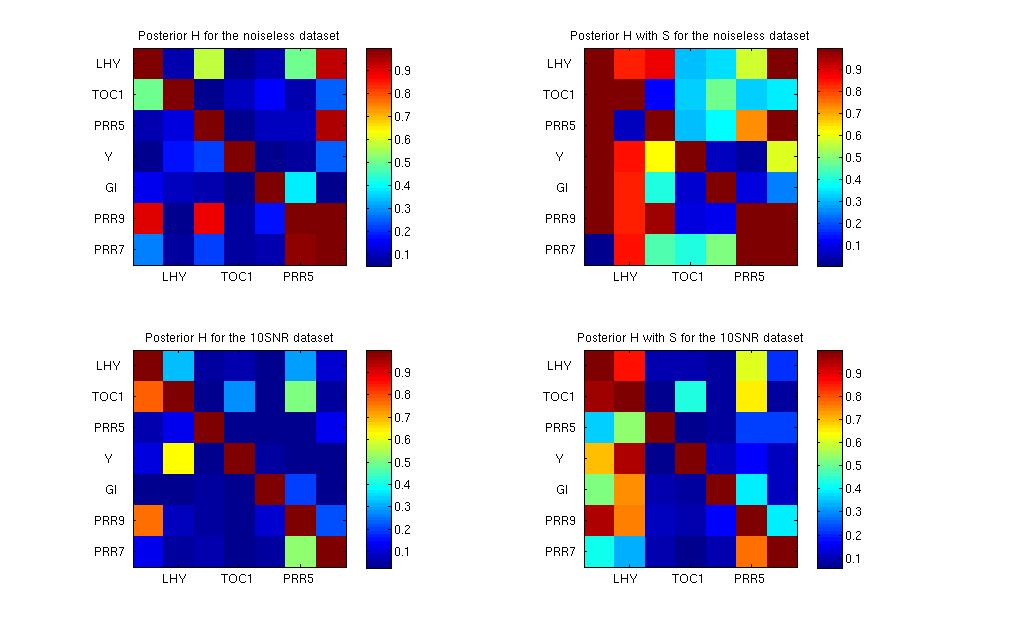}\caption{Heatmaps representing the posterior probability for the \emph{A. thaliana
}circadian clock network\label{fig:HeatmapsCircadian}}

\end{figure}

\begin{figure}[H]
\includegraphics[width=1\textwidth]{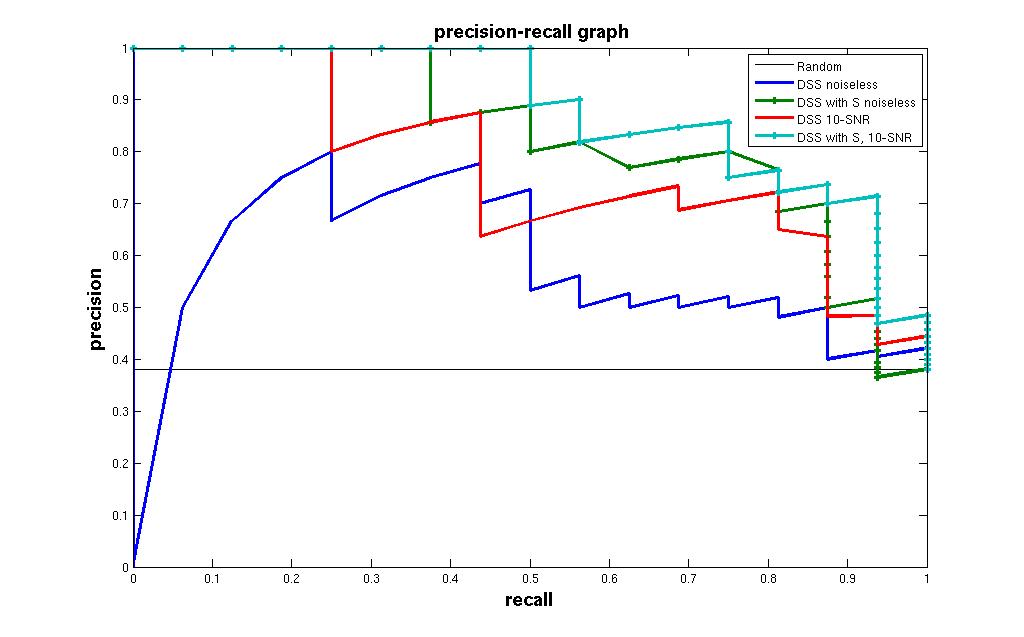}\caption{PR curves for the \emph{A. thaliana }circadian clock network\label{fig:PRcircadian}}
\end{figure}

\begin{figure}[H]
\includegraphics[width=1\textwidth]{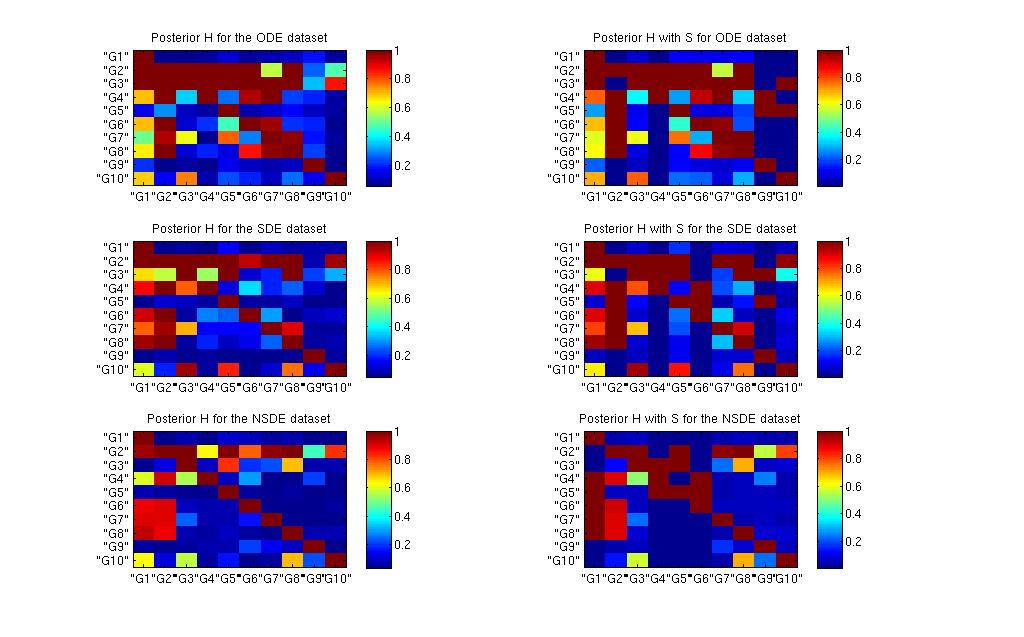}\caption{Heatmaps representing the posterior probability for the Dream4 Challenge
network\label{fig:HeatmapsDream}}
\end{figure}

\begin{figure}[H]
\includegraphics[width=1\textwidth]{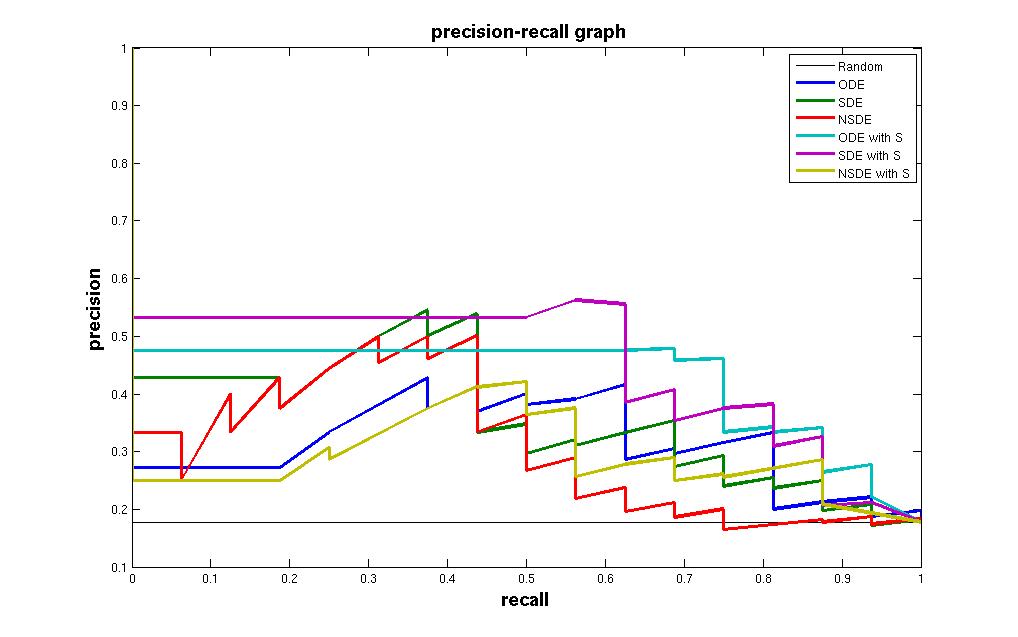}\caption{PR curves for the DREAM4 challenge network\label{fig:PRDream}}
\end{figure}

\begin{figure}[H]
\includegraphics[width=1\textwidth]{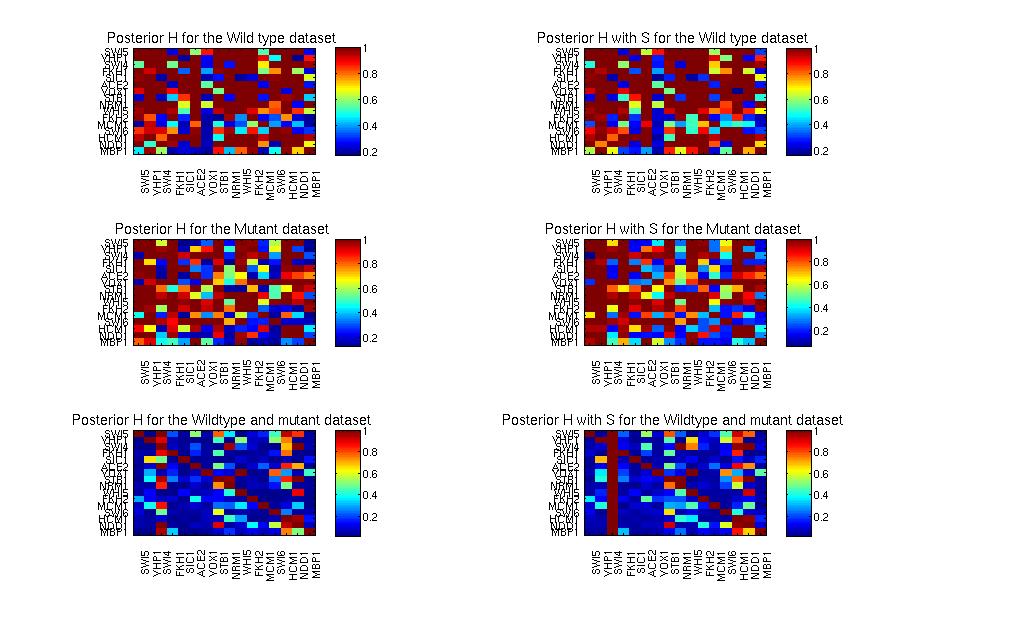}\caption{Heatmaps representing the posterior probability for the \emph{s. cerevisiae
}network\label{fig:HeatmapsCircadian-2}}
\end{figure}

\begin{figure}[H]
\includegraphics[width=1\textwidth]{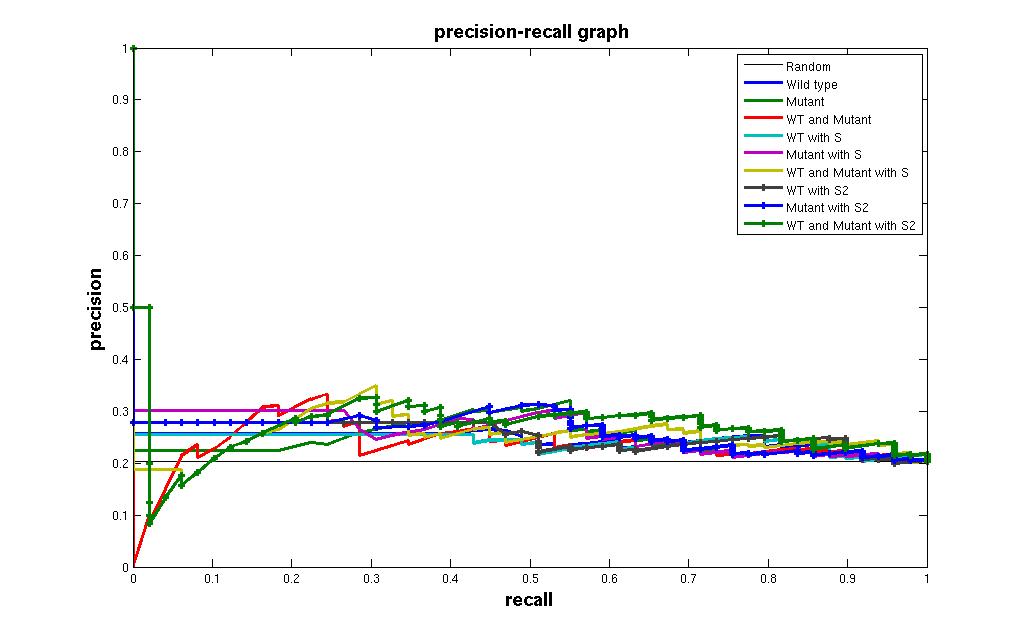}\caption{PR curves for the \emph{s. cerevisiae }network\label{fig:PRcircadian-2}}
\end{figure}

\section{\emph{A. thaliana }circadian clock}

\begin{figure}[h]
\begin{centering}
\includegraphics[width=0.7\textwidth]{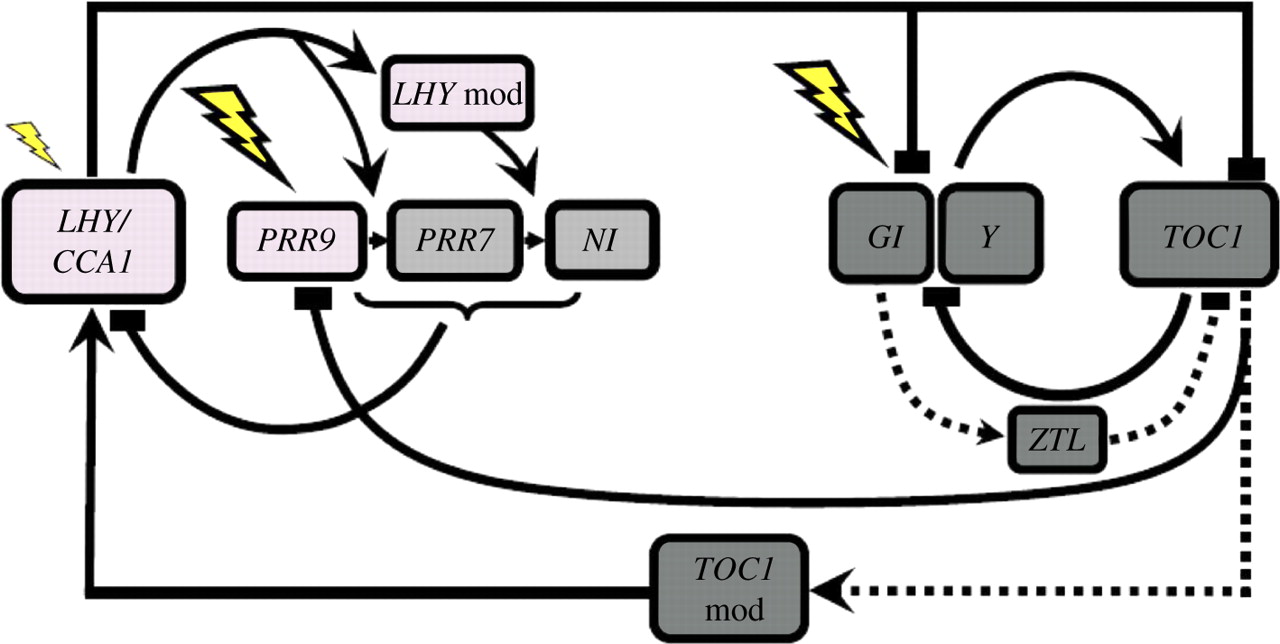}
\par\end{centering}

\caption{\emph{A. thaliana }circadian clock model, transcriptional elements
LHY, PRR9, PRR7, NI, Y, and TOC1. Post-transcriptional elements ZTL,
TOC1mod and LHYmod. Light input is represented by a lighting symbol.
Activating interactions are represented by solid line with arrows,
repression by solid line with rectangles at the end, post transcriptional
interactions are represented by dashed lines. \label{fig:Pokhilkho}.}
\end{figure}

The arabidopsis thaliana circadian clock model as presented in, is
shown in Fig. \ref{fig:Pokhilkho}. 

\section{DREAM4 network}

\begin{figure}[H]
\begin{centering}
\includegraphics[width=0.8\textwidth]{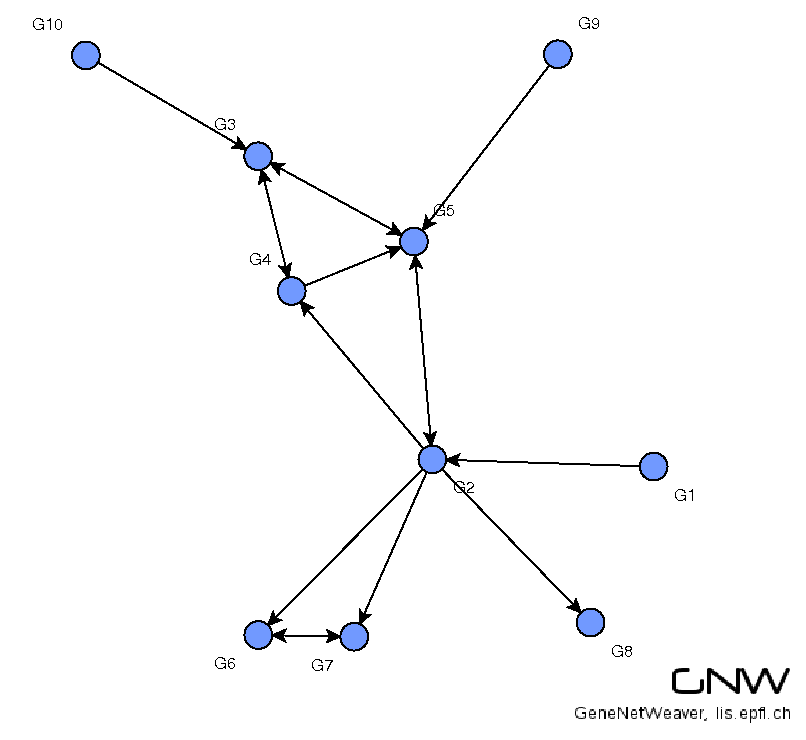}
\par\end{centering}

\caption{DREAM4 challenge network with 10 nodes, of those 3 are inputs, node
G9 was subjected to a perturbation for half the time points\label{fig:DREAM4-challenge-network}.}
\end{figure}

The 10-node oscillatory network that was part of the DREAM4 challenge
supplementary information data set is presented in Fig. \ref{fig:DREAM4-challenge-network}.

\section{\emph{S. cerevisae }cell cycle network}

\begin{figure}[H]
\begin{centering}
\includegraphics[width=1\textwidth]{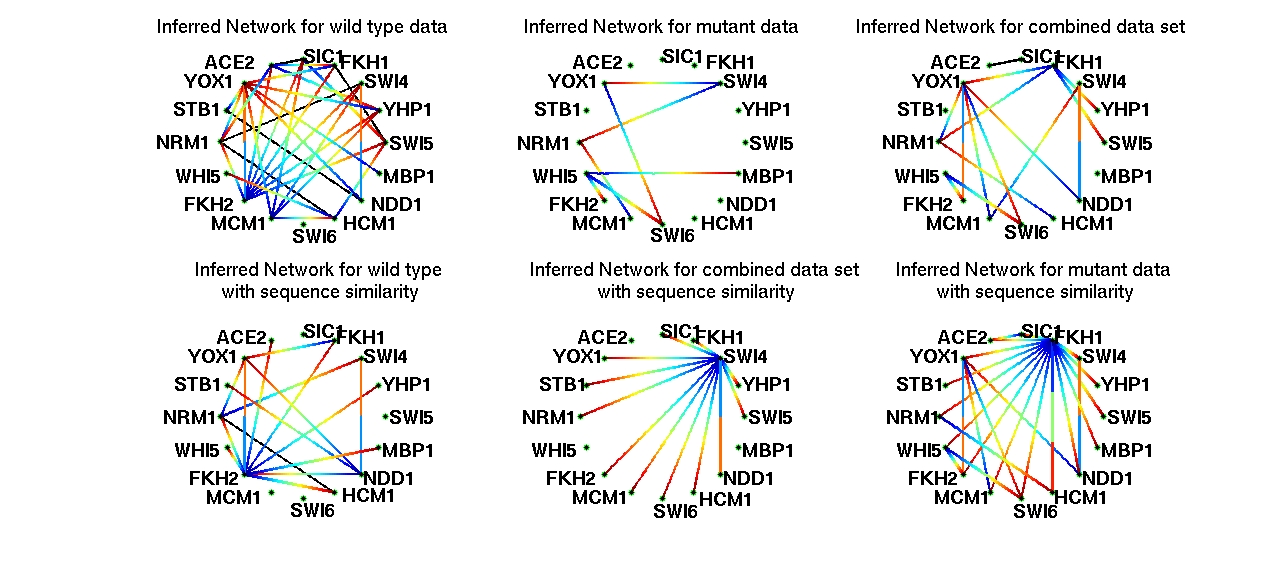}
\par\end{centering}

\caption{Inferred yeast networks for different data subsets with and without
sequence information, edges go from blue (regulators) to red (targets).\label{yeast nets}.}
\end{figure}

\begin{figure}[H]
\begin{centering}
\includegraphics[width=1\textwidth]{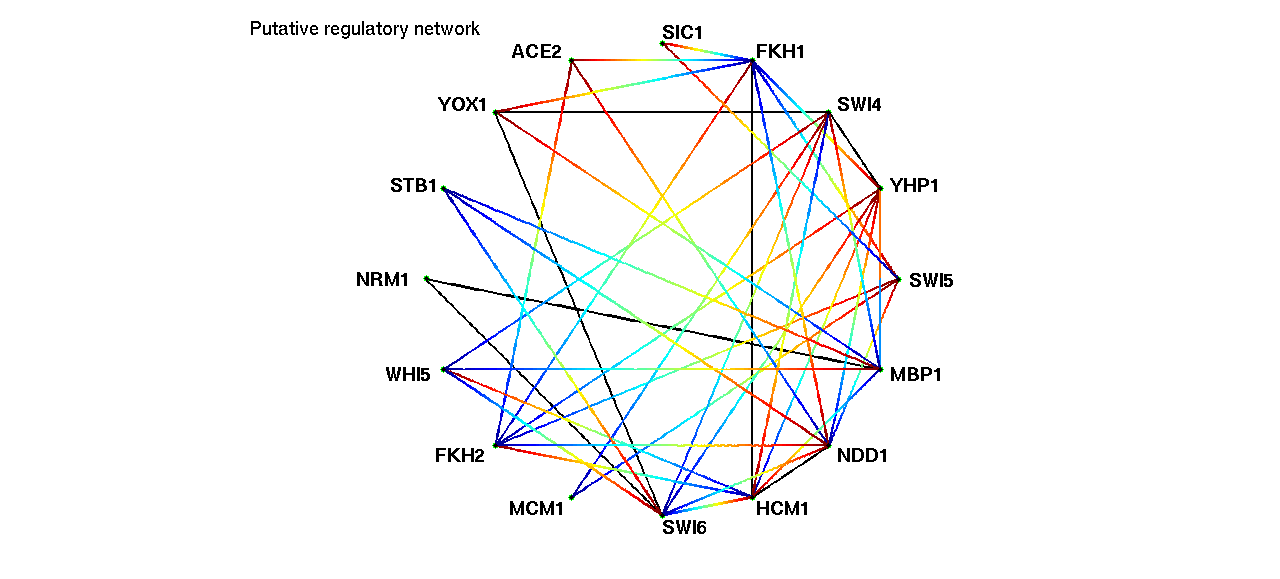}
\par\end{centering}

\caption{Putative yeast regulatory network edges go from blue (regulators)
to red (targets).\label{yeastputative}.}
\end{figure}

In Fig. \ref{yeast nets} the inferred networks after thresholding
the value of $p\left(h_{ij}=1\right)$ are presented, the putative
ground truth matrix is presented on Fig. \ref{yeastputative}

The putative regulatory network was built based on these references. \begin{itemize} \item Regulation of SIC1 by SWI5 as in \cite{Weiss_2012}.  \item Regulation of SWI4 by YHP1  as in \cite{Bahler_2005}. \item Regulation of YHP1, SWI4, YOX1 and HCM1 by SWI4 as in \cite{Macisaac_2006}. \item Regulation of SWI5 and ACE2 \cite{Haase_2014}; YHP1 \cite{Macisaac_2006}; SIC1, YOX1 and HCM1 \cite{venters_2011}; NDD1 \cite{Ostrow_2014}; by FKH1. \item Regulation of SWI6 and MBP1 by NRM1 as in \cite{Macisaac_2006}. \item Regulation of SWI6, SWI4 and MBP1 by WHI5 as in \cite{Haase_2014}. \item Regulation of SWI5, YHP1 and FKH1 \cite{Macisaac_2006}; ACE2 and NDD1 \cite{Haase_2014} by FKH2. \item Regulation of SWI4 and SWI5 by MCM1 as in \cite{Macisaac_2006}. \item Regulation of SWI4, FKH1, YOX1, NRM1, HCM1 and NDD1 as in \cite{Macisaac_2006}. \item Regulation of YHP1, FKH1, FKH2, WHI5 and NDD1 by HCM1 as in \cite{Pramila_2006}. \item Regulation of SWI5 and ACE2 \cite{Haase_2014}; YHP1 and HCM1 as in \cite{Macisaac_2006}. \item Regulation of YHP1, FKH1, YOX1, NRM1 and HCM1 by MBP1 as in \cite{Macisaac_2006}. \item Regulation of NDD1 \cite{Macisaac_2006}; SWI6 and MBP1 by the interaction of transcription factor MBF with STB1 \cite{stillman_2013}. \item Regulation of SWI4 and SWI6 by its cobinding with MCM1p as in \cite{Haase_2014}. \end{itemize}
 \end{document}